\documentclass{article}

\usepackage{PRIMEarxiv}

\usepackage[utf8]{inputenc}
\usepackage[T1]{fontenc}
\usepackage{xr-hyper}
\usepackage{hyperref}
\usepackage{url}
\usepackage{booktabs}
\usepackage{amsfonts}
\usepackage{nicefrac}
\usepackage{microtype}
\usepackage{fancyhdr}
\usepackage{graphicx}
\graphicspath{{media/}}
\usepackage{amsmath}
\usepackage{amssymb}
\usepackage{xcolor}
\usepackage{placeins}
\usepackage{subcaption}
\title{Learning atomic interactions for the thermodynamics of liquid mixtures}

\author{
  Marco Hoffmann\\
  RPTU Kaiserslautern-Landau\\
  Kaiserslautern\\
  \texttt{marco.hoffmann@rptu.de} \\
   \And
  Sebastian Schmitt\\
  RPTU Kaiserslautern-Landau\\
  Kaiserslautern\\
  \texttt{sebastian.schmitt@rptu.de} \\
   \And
  Hans Hasse\\
  RPTU Kaiserslautern-Landau\\
  Kaiserslautern\\
  \texttt{hans.hasse@rptu.de} \\
  \And
  Fabian Jirasek\\
  RPTU Kaiserslautern-Landau\\
  Kaiserslautern\\
  \texttt{fabian.jirasek@rptu.de} \\
}

\begin{document}
\maketitle

\begin{abstract}
Predictive models for activity coefficients, which describe the non-ideality of liquid mixtures, are foundational for process design and optimization. Group-contribution methods, such as UNIFAC, are computationally cheap and well established in process simulators. Modern machine learning models, such as HANNA, offer higher accuracy and broader applicability, but their integration into process simulation is expensive or even infeasible. In this work, we introduce COGNAC, a graph neural network that learns atom-level geometry and interaction parameters that depend on the molecular neighborhood and evaluates them in the UNIFAC framework. Trained on about 820,000 experimental data points, COGNAC achieves significantly higher accuracy than UNIFAC and HANNA for binary and multi-component mixtures, reduces error relative to HANNA by 13--63\%, reliably predicts multi-phase equilibria and identifies azeotropes, yields interpretable parameters, and is strictly thermodynamically consistent. COGNAC evaluations run at a cost comparable to UNIFAC, thereby unlocking the potential of machine-learning models for process simulations.
\end{abstract}

\section{Main}
The feasibility and cost of distillation, extraction, and crystallization processes are governed by the non-ideality of the liquid phase. This non-ideality is quantified by the activity coefficients of the components, which are therefore central to the design and optimization of chemical processes. Because experimental data are unavailable for most mixtures of interest, activity coefficients are predicted with models for the excess Gibbs energy $g^\mathrm{E}$, which express the non-ideality as a function of composition and temperature.

The most established predictive excess Gibbs energy models are group contribution (GC) methods, in particular UNIFAC and its variants \cite{Fredenslund1975, Magnussen1981, Weidlich1987, Wittig2002, Lei2009, Constantinescu2016}. They decompose molecules into predefined structural groups and describe non-ideal mixture behavior through interactions between these groups. These models have been widely adopted in chemical engineering because they are computationally inexpensive and are implemented in most process simulators. At the same time, the predefined group structure limits the molecular resolution of UNIFAC models because they resolve the molecular neighborhood only coarsely. For example, modified UNIFAC (Dortmund), called mod.~UNIFAC in the following, accounts for the molecular neighborhood of a hydroxyl group using subgroups that distinguish only the immediate bonding situation (OH bound to a primary, secondary, or tertiary carbon), but share their interaction parameters such that effects that extend beyond the nearest neighbors are lost. This limits the description of multifunctional molecules and structural isomers that decompose into the same groups. Furthermore, GC methods apply only to decomposable components for which both suitable structural groups and fitted parameters are available.

Quantum-chemistry (QC) based methods, such as COSMO-RS~\cite{Klamt1995} and COSMO-SAC~\cite{Lin2001}, compute activity coefficients from molecular surface-charge distributions and need no component-specific parameterizations. However, they require an expensive QC calculation per component and have not shown a systematic accuracy advantage over GC methods where both apply~\cite{Grensemann2005,Fingerhut2017, Xue2012}.

machine learning (ML) has recently opened alternatives to predictive thermodynamics, including hybrid strategies~\cite{Jirasek2023, Jirasek2020b, Hasse2026} that embed ML in an established physics-based model and train it to predict the model's parameters~\cite{Hoffmann2024, Hoffmann2025, Jirasek2022, Jirasek2023a}. Hayer et al.~\cite{Hayer2025,Hayer2025a} applied matrix completion methods (MCM) to complete the interaction parameter tables of UNIFAC~\cite{Fredenslund1975} and mod.~UNIFAC~\cite{Weidlich1987}, yielding models with broader applicability and higher accuracy than their predecessors. However, these new methods still rely on the predefined functional groups of the original UNIFAC formulation, and certain types of molecules (e.g., most ionic liquids) remain out of scope. Following a similar idea, SPT-NRTL~\cite{Winter2023} predicts the parameters of the Non-Random-Two-Liquid (NRTL) model~\cite{Renon1968} from SMILES with a transformer~\cite{Winter2022}.

Another hybrid strategy uses flexible neural networks to directly predict excess Gibbs energy or activity coefficients while incorporating physical constraints into the ML architecture or training loss~\cite{SanchezMedina2023, Rittig2023, Rittig2024, Specht2024, Hoffmann2026}. The most accurate of these models, HANNA~\cite{Specht2024,Hoffmann2026}, relies on pre-trained molecular embeddings (ChemBERTa-2~\cite{Ahmad2022}) and enforces Gibbs--Duhem consistency, equivariance, and the pure-component limits by construction. The hybrid models are accurate and broadly applicable because they need only molecular structure as input. However, because the neural network is part of the thermodynamic model, it must be evaluated at every simulation step. This makes neural-network-based models prohibitively expensive for simulation purposes.

Here, we introduce COGNAC (local-\textbf{CO}mposition \textbf{G}raph \textbf{N}eural network for the prediction of \textbf{A}ctivity \textbf{C}oefficients), a $g^\mathrm{E}$ model that combines a graph neural network (GNN) with the local-composition framework of mod.~UNIFAC. In this framework, mixture non-ideality arises from an ensemble of pairwise interacting sites, a picture that QC-based methods also share (Supplementary Note~\ref{SI-si:note_gc_qc}). COGNAC retains the local-composition framework but fundamentally changes what an interaction site is: instead of predefined structural groups, it uses individual atoms, whose geometry and interaction parameters depend on their molecular neighborhood. In COGNAC, each atom's molecular neighborhood extends over several bonds and influences all its parameters, including pair interactions. COGNAC therefore captures how neighboring functional groups influence each other, something structural groups cannot resolve. For the user, a single network evaluation turns the SMILES of the components in a mixture into a mod.~UNIFAC parameter table that any process simulator can use.

\begin{figure}[!h]
    \centering
    \includegraphics[width=\linewidth]{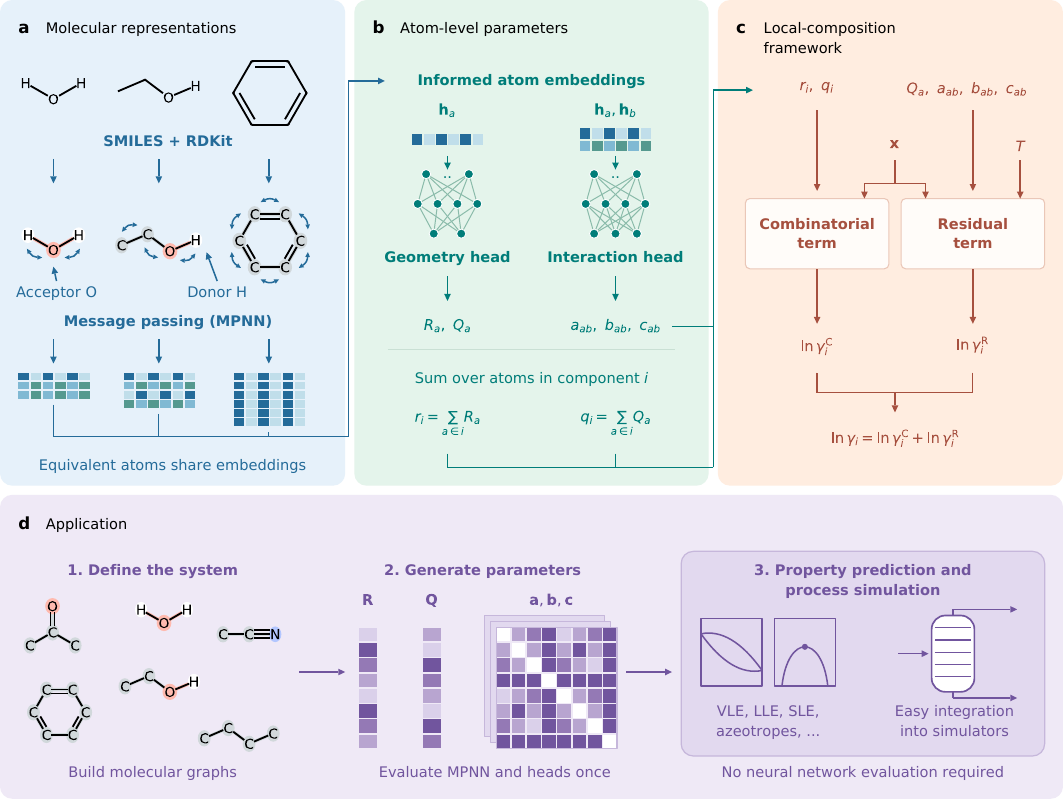}
    \begin{subfigure}[b]{0pt}\phantomsubcaption\label{fig:cognac_overview_molecular_representation}\end{subfigure}%
    \begin{subfigure}[b]{0pt}\phantomsubcaption\label{fig:cognac_overview_atom_level_parameters}\end{subfigure}%
    \begin{subfigure}[b]{0pt}\phantomsubcaption\label{fig:cognac_overview_atom_local_composition}\end{subfigure}%
    \begin{subfigure}[b]{0pt}\phantomsubcaption\label{fig:cognac_overview_atom_application}\end{subfigure}%
    \caption{Overview of the COGNAC model architecture and its application.
    \textbf{a}, Molecular graphs constructed from SMILES~\cite{Weininger1988} using RDKit~\cite{rdkit} are encoded by a message-passing neural network (MPNN). Heavy atoms and hydrogens bound to non-carbon atoms form explicit nodes, allowing hydrogen-bond donors and acceptors to be resolved. Atoms with identical molecular neighborhoods within the message-passing radius obtain identical embeddings.
    \textbf{b}, Two prediction heads map the context-aware atom embeddings $\mathbf{h}_a$ to volume and surface-area parameters $R_a$ and $Q_a$, and pairs of embeddings to the symmetric interaction coefficients $a_{ab}$, $b_{ab}$, and $c_{ab}$ of the exported parameter table (Methods). Summing the respective atom-level geometry parameters gives the molecular quantities $r_i$ and $q_i$.
    \textbf{c}, The learned parameters and the state point (composition $\mathbf{x}$ and temperature~$T$) are processed in a local-composition framework. The combinatorial contribution accounts for size and shape effects, while atomic surface areas and pair interactions determine the residual contribution. Together, the two contributions yield the activity coefficient of component $i$.
    \textbf{d}, For a specified component set, parameters need to be generated only once and can be reused across compositions and temperatures without further neural network evaluations. The obtained atom-level parameter vectors $\mathbf{R}$ and $\mathbf{Q}$ and interaction matrices $\mathbf{a}$, $\mathbf{b}$, and $\mathbf{c}$ enable, e.g., the prediction of vapor--liquid (VLE), liquid--liquid (LLE), and solid--liquid (SLE) equilibria, azeotropes, and residue curve maps for binary and multi-component mixtures, and can easily be implemented in process simulations.
    }
    \label{fig:cognac_overview}
\end{figure}

Fig.~\ref{fig:cognac_overview} gives an overview of the COGNAC architecture. A message-passing neural network~\cite{Gilmer2017} encodes each atom in each component together with its molecular neighborhood within four bonds; symmetry-equivalent atoms receive identical embeddings, and hydrogens bound to non-carbon atoms are explicit nodes, making hydrogen-bond donors and acceptors distinct sites. From these embeddings, one network predicts each atom's volume and surface-area parameters, and another predicts the interaction parameters for every atom pair. The parameters enter the unchanged mod.~UNIFAC equations: molecular geometry in the combinatorial contribution, surface areas and pair interactions in the residual contribution.

COGNAC is trained end-to-end on 822,780 experimental data points for 48,432 binary mixtures from the Dortmund Data Bank (DDB)~\cite{DDB2026}, including vapor--liquid equilibria (VLE), liquid--liquid equilibria (LLE), solid--liquid equilibria (SLE), activity coefficients at infinite dilution (ACT), and excess enthalpies (HE). The learned atomic surface areas and molecular volumes are regularized against COSMO~\cite{Klamt1993} cavity data from the CHAOS database~\cite{Gond2026}. Training on LLE data is enabled by a differentiable phase-split solver combining the convex envelope method~\cite{Ryll2012, Goettl2023, Goettl2025} with a Newton step (Methods).

Separating parameter prediction from thermodynamic evaluation has several advantages. The excess Gibbs energy and all derived properties are smooth and satisfy the Gibbs--Duhem equation, permutation symmetry, and the pure-component limits by construction; the same equations describe binary and multi-component mixtures without additional parameters or a separate multi-component approximation; the learned geometry and interaction parameters are interpretable; and because they are generated once for a specified component set and reused across compositions and temperatures, phase-equilibrium calculations and process simulations require no further neural network evaluations. 

\clearpage
\section{Results}

\subsection{Prediction of phase equilibria and excess properties}
\begin{figure}[h]
    \centering
    \includegraphics[width=\linewidth]{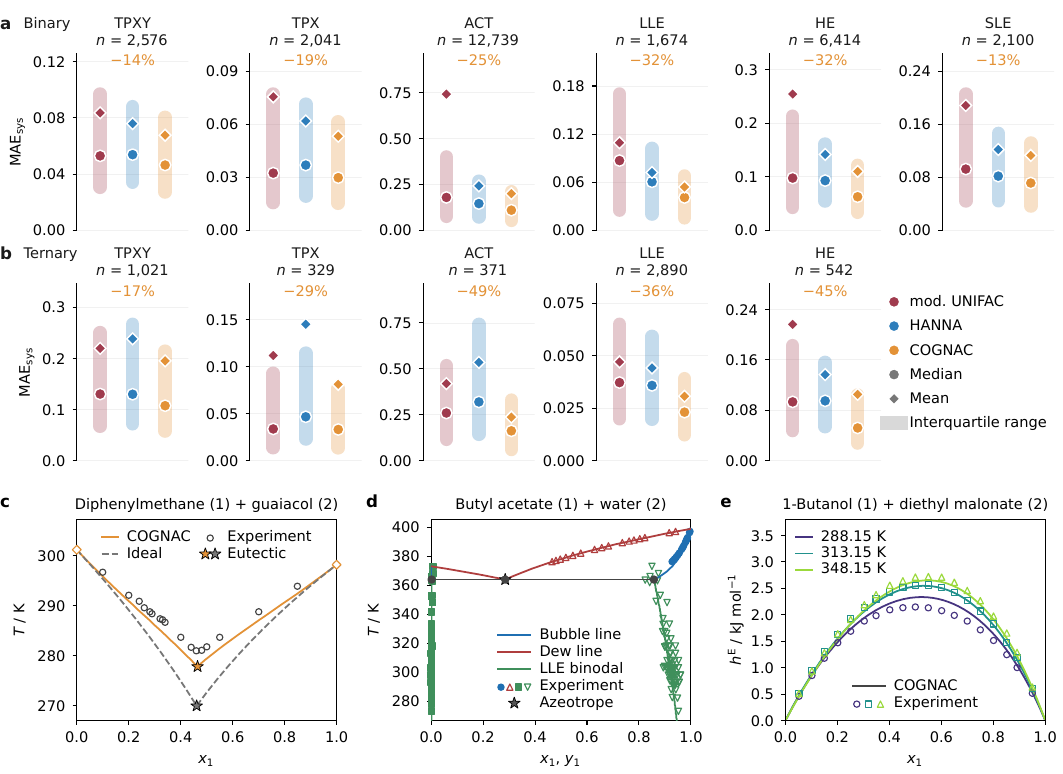}%
    \begin{subfigure}[b]{0pt}\phantomsubcaption\label{fig:binary_boxplot}\end{subfigure}%
    \begin{subfigure}[b]{0pt}\phantomsubcaption\label{fig:ternary_boxplot}\end{subfigure}%
    \caption{Results of COGNAC and comparison to mod.~UNIFAC and HANNA. \textbf{a}, System-wise mean absolute errors $\mathrm{MAE}_\mathrm{sys}$ for the binary data, resolved by data type. Results of COGNAC and HANNA are obtained by system-wise ten-fold cross-validation, ensuring unseen test systems. Data types: activity coefficients from vapor--liquid equilibria with (TPXY) and total pressures without (TPX) reported vapor composition, activity coefficients at infinite dilution (ACT), liquid--liquid equilibria (LLE), excess enthalpies (HE), and activity coefficients from solid--liquid equilibria (SLE). The scores are calculated from the absolute deviations in the logarithmic activity coefficient (TPXY, ACT, SLE), logarithmic pressure (TPX), liquid mole fractions (LLE), and excess enthalpies in kJ\,mol$^{-1}$ (HE). Circles denote the median, diamonds the mean, and the shaded bars the interquartile range of the distribution of $\mathrm{MAE}_\mathrm{sys}$ over all systems; $n$ is the number of systems in the respective panel, restricted to those systems that all three models can describe. The percentage above each panel shows the relative change in the median $\mathrm{MAE}_\mathrm{sys}$ of COGNAC relative to HANNA; negative values indicate that COGNAC is more accurate. Mod.~UNIFAC describes 50\% of the binary systems of the full data set (Extended Data Fig.~\ref{ed:fig_benchmark_all}). COGNAC and HANNA were trained and evaluated on the same data, mod.~UNIFAC was used as published. \textbf{b}, Same comparison for the ternary data set, predicted with the full-data models of COGNAC and HANNA.  Mod.~UNIFAC describes 52\% of the ternary systems of the full data set. \textbf{c}--\textbf{e}, Exemplary COGNAC predictions for different binary data types against experimental data from unseen systems: solid--liquid equilibrium with eutectic point (\textbf{c}; the dashed line is the ideal-solution liquidus), heteroazeotrope with three-phase equilibrium at atmospheric pressure (\textbf{d}; the horizontal line marks the temperature at which vapor and two liquid phases coexist; below it, the green lines are the compositions of the two coexisting liquid phases), and excess enthalpies at different temperatures (\textbf{e}).}
    \label{fig:benchmark_boxplots}
\end{figure}

We benchmark COGNAC against mod.~UNIFAC~\cite{Constantinescu2016}, the most established predictive activity coefficient model, and HANNA~\cite{Hoffmann2026}, the most accurate published ML model, on binary and ternary mixtures (Fig.~\ref{fig:benchmark_boxplots}). 

We compare six data types: activity coefficients from VLE with specified vapor composition (TPXY), total pressures from VLE without reported vapor composition (TPX), phase compositions in LLE, activity coefficients derived from SLE, data for activity coefficients at infinite dilution (ACT), and excess enthalpies (HE). For each data type, we calculate the system-wise mean absolute errors $\mathrm{MAE}_\mathrm{sys}$, which we use as the primary error metric (Methods). Binary results are out-of-fold predictions of a ten-fold cross-validation over systems with HANNA retrained on identical folds; ternary results use COGNAC and HANNA trained once on all binary data; no multi-component data was used for training. Mod.~UNIFAC, used as published, describes only 43--83\% of the binary and 21--85\% of the ternary systems depending on the data type. Fig.~\ref{fig:benchmark_boxplots} therefore compares the systems all three models describe; the full data set comparison between COGNAC and HANNA is shown in Extended Data Fig.~\ref{ed:fig_benchmark_all}.

For the binary mixtures, COGNAC yields the lowest median and mean $\mathrm{MAE}_\mathrm{sys}$ across all data types. Relative to HANNA, the median error is 13--32\% lower across data types, and relative to mod.~UNIFAC, 8--53\% lower. The improvement is consistent: HANNA is ahead only in one fold on one data type (SLE), so no data type improves at the expense of another. This advantage of COGNAC holds also on the full data set, including the systems mod.~UNIFAC cannot describe (Extended Data Fig.~\ref{ed:fig_benchmark_all}).

HANNA extrapolates from binary to ternary mixtures via geometric projection, whereas COGNAC evaluates the same equations over atoms from any number of components. This mechanistic extrapolation pays off (Fig.~\ref{fig:ternary_boxplot}): COGNAC reduces the median $\mathrm{MAE}_\mathrm{sys}$ relative to HANNA by 17--49\%, more than for the binary mixtures for every data type and most clearly for ACT (49 vs.\ 25\%) and HE (45 vs.\ 32\%). It is also at least as accurate as mod.~UNIFAC for every ternary data type (equal medians for TPX, 18--44\% lower otherwise), although mod.~UNIFAC was fitted to ternary data and COGNAC was not. The same holds for quaternary mixtures (Extended Data Fig.~\ref{ed:fig_quaternary}): 27--63\% lower median errors than HANNA and 9--18\% lower than mod.~UNIFAC on the systems all three models describe, and 22--61\% lower than HANNA on the full data set.

This advantage holds across architectural choices: in an ablation study of the temperature dependence of the interaction parameters and the residual-term closure, all six variants beat HANNA on every binary data type (Extended Data Table~\ref{ed:tab_ablation_study}, Supplementary Note~\ref{SI-si:note_ablation}). COGNAC uses the closed-form residual term of UNIFAC, which sets the surrounding sites to unit activity, an approximation to COSMOSPACE's self-consistent closure~\cite {Klamt2002} that must be solved iteratively. We also tested a self-consistent variant, COGNAC-IMP, that improves the mean $\mathrm{MAE}_\mathrm{sys}$ of binary activity coefficients at infinite dilution by 4.5\% and of excess enthalpies by 4.8\%, with mixed effects on the other data types and ternary and quaternary mixtures (Supplementary Note~\ref{SI-si:note_imp}, Supplementary Table~\ref{SI-si:tab_imp_vs_exp}). We therefore use the closed form for COGNAC, which keeps the mod.~UNIFAC equations unchanged and needs no iterative solver.

Fig.~\ref{fig:benchmark_boxplots}c--e shows three out-of-fold predictions: a binary SLE, a heteroazeotrope, and excess enthalpies at three temperatures. The heteroazeotrope in Fig.~\ref{fig:benchmark_boxplots}d addresses a long-standing problem: classical $g^\mathrm{E}$ models~\cite{Rarey2005, Marcilla2016} struggle to describe VLE and LLE simultaneously with a single parameterization, and the original UNIFAC required a separate LLE parameter table~\cite{Magnussen1981}. Accordingly, mod.~UNIFAC is about as accurate as HANNA for VLE but considerably less accurate for LLE (Fig.~\ref{fig:binary_boxplot}), whereas HANNA, due to its greater flexibility, describes both well~\cite{Hoffmann2026}. COGNAC is the most accurate for both using the unchanged mod.~UNIFAC equations. This suggests that UNIFAC's limitation stems not from its equations but from its rigid group definitions. 

Predicting the behavior of aqueous mixtures remains challenging for thermodynamic models: water is a small, strongly self-associating molecule, and hydrogen bonding governs mixture behavior. Aqueous mixtures are the hardest case for all models (Extended Data Fig.~\ref{ed:fig_aqueous}), yet COGNAC is more accurate than HANNA for every data type here as well, with the largest gains for ternary VLE and ACT. We attribute this to the explicit polar hydrogen nodes: water is represented by three interaction sites, and a hydrogen bond is a donor--acceptor pair interaction rather than a collapsed property of the water molecule. In an ablation that treats all hydrogen atoms implicitly, with architecture, data, hyperparameters, and seed unchanged (which also removes the donor--acceptor sign regularizer, Methods), the median error of aqueous systems rises by 31\% but that of non-aqueous systems by only 1.7\%, and aqueous mixtures get worse in all eleven combinations of mixture order and data type (Extended Data Fig.~\ref{ed:hydrogen_ablation}).

A known weakness of mod.~UNIFAC is largely removed by COGNAC: with its published parameters, mod.~UNIFAC underestimates the activity coefficients of alkanes at infinite dilution in water, a compromise that was deliberately accepted in favor of an accurate description of alcohol--water systems~\cite{Jakob2006}. COGNAC, evaluating the same equations, describes alkanes and alcohols in water well (Extended Data Fig.~\ref{ed:fig_aqueous_solute_parity}).

COGNAC also resolves structural isomers with identical group counts, for which mod.~UNIFAC predicts the same activity coefficients by construction. Among such isomer pairs whose measured $\ln \gamma^\infty$ differ by at least 0.15, COGNAC reproduces the experimental order in 84.5\% of 1,368 cases (Extended Data Fig.~\ref{ed:fig_degeneracy}, Supplementary Note~\ref{SI-si:note_degeneracy}). 

\FloatBarrier
\subsection{Prediction of azeotropes and residue curve maps}
\label{sec:azeotropes}

\begin{figure}[!b]
    \centering
    \includegraphics[width=\linewidth]{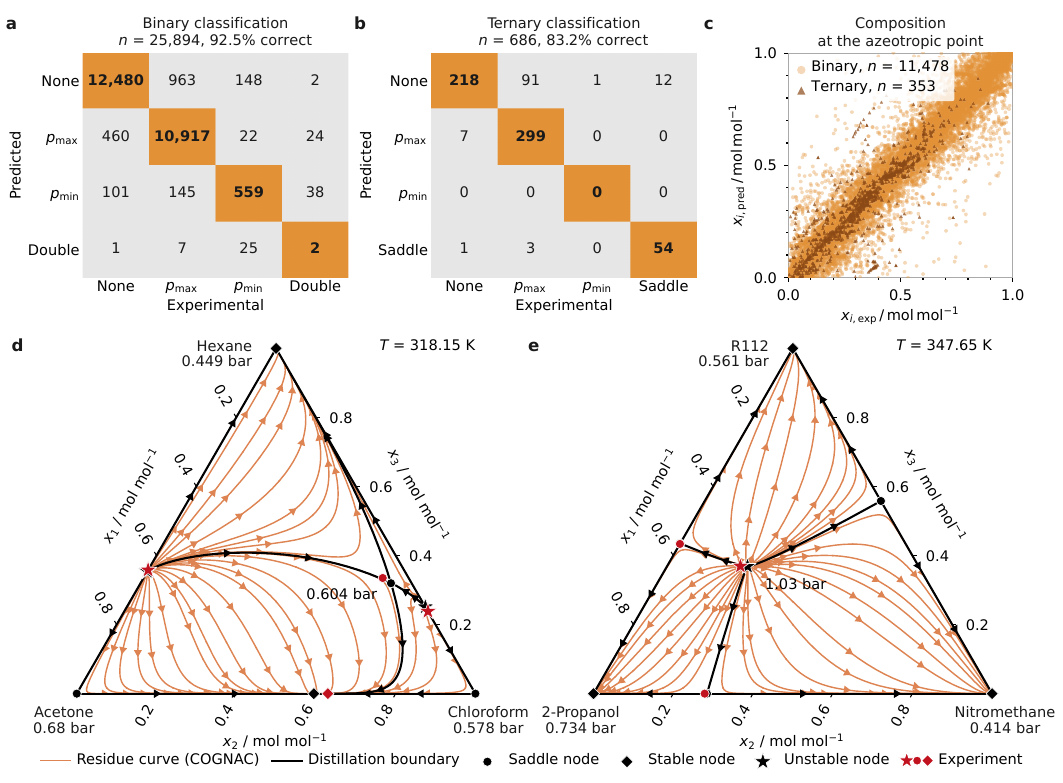}%
    \begin{subfigure}[b]{0pt}\phantomsubcaption\label{fig:binary_azeotrope}\end{subfigure}%
    \begin{subfigure}[b]{0pt}\phantomsubcaption\label{fig:ternary_azeotrope}\end{subfigure}%
    \begin{subfigure}[b]{0pt}\phantomsubcaption\label{fig:azeotrope_composition}\end{subfigure}%
    \begin{subfigure}[b]{0pt}\phantomsubcaption\label{fig:rcm_1}\end{subfigure}%
    \begin{subfigure}[b]{0pt}\phantomsubcaption\label{fig:rcm_2}\end{subfigure}%
    \caption{COGNAC predictions of azeotropes and residue curve maps. \textbf{a,b}, Classification matrices for azeotropes in binary (\textbf{a}) and ternary (\textbf{b}) mixtures: predicted class against experimental class. The diagonal marks correct prediction of azeotrope existence and class with COGNAC; the corresponding matrices for HANNA and mod.~UNIFAC are shown in Extended Data Fig.~\ref{ed:fig_azeotrope_models}. Azeotrope types are minimum-boiling ($p_\mathrm{max}$), maximum-boiling ($p_\mathrm{min}$), double azeotrope (double), saddle-point (saddle), and zeotropic (none). $n$ denotes the number of data points. \textbf{c}, Predicted compositions at the azeotropic point, compared with experimental values, for all components in the mixture. Only azeotropes for which existence and type were predicted correctly are shown; $n$ is the number of azeotropes, each contributing one point per component. \textbf{d,e}, Isothermal residue curve maps predicted by COGNAC for acetone + chloroform + hexane at 318.15\,K (\textbf{d}) and for 2-propanol + nitromethane +  1,1,2,2-tetrachloro-1,2-difluoroethane (R112) at 347.65\,K (\textbf{e}), a mixture outside the scope of mod.~UNIFAC (no group decomposition for R112). Orange lines are residue curves, black lines are the distillation boundaries. The pressures at the vertices are the pure-component vapor pressures at the given temperature; the pressure next to a ternary azeotrope is its predicted value. Predicted singular points are marked by symbols encoding their type (unstable node, stable node, saddle). Experimental azeotropes are shown in red: the ternary azeotrope with its reported type, binary azeotropes measured within 5\,K with the symbol of the nearest predicted singular point on that edge.}
    \label{fig:azeotropes}
\end{figure}

Azeotropes, where vapor and liquid have the same composition, limit the feasibility of simple distillation and, in multi-component mixtures, divide the composition space into distillation regions that residue curves cannot cross. We assess COGNAC's performance in predicting azeotropes on the DDB azeotropic data bank~\cite{DDB2026}, which contains both positive samples (azeotropic mixtures and their azeotrope type) and negative samples (zeotropic mixtures). We locate all homogeneous azeotropes from
\begin{equation}
\label{eq:azeotrope_condition}
y_i(\mathbf{x},T) = x_i, \qquad i = 1,\dots,N-1,
\end{equation}
at the reported temperature (Supplementary Note~\ref{SI-si:note_azeotrope_methods}). A prediction is counted as correct if it reproduces the existence and type of the measured azeotrope. Binary azeotropes are out-of-fold predictions; ternary azeotropes are predicted with the full-data models, trained on all binary data. Fig.~\ref{fig:azeotropes} summarizes the results.

Figs.~\ref{fig:binary_azeotrope} and \ref{fig:ternary_azeotrope} compare the predicted azeotrope type with the measured one. Binary azeotropes are minimum-boiling ($p_\mathrm{max}$) or maximum-boiling ($p_\mathrm{min}$) azeotropes, and a few binary mixtures show both at the same temperature (double). Ternary azeotropes can additionally form saddle-point azeotropes. COGNAC correctly classifies $92.5\%$ of the 25,894 data points and HANNA $90.9\%$; mod.~UNIFAC applies to only $76\%$ of them, where it correctly classifies $92.1\%$ against $92.8\%$ for COGNAC on the same subset (Extended Data Fig.~\ref{ed:fig_azeotrope_models}). For the 686 ternary data points, COGNAC reaches 83.2\% and HANNA 64.7\% accuracy; mod.~UNIFAC applies to 83\% and classifies 78.5\% of those correctly, against 85.5\% for COGNAC. The three models are thus similar for binary azeotropes, and COGNAC is clearly ahead for ternary ones. The one class COGNAC misses is the rare double azeotrope: it recovers 2 of the 58 measured points that all three models evaluate, HANNA 36 and mod.~UNIFAC 28 (Supplementary Table~\ref{SI-si:tab_azeotrope_classes}).

Fig.~\ref{fig:azeotrope_composition} shows that COGNAC also predicts the azeotropic compositions well. For the azeotropes that all three models classify correctly, the median absolute deviation over all the mole fractions is $0.020$ (binary) and $0.022\,\mathrm{mol}\,\mathrm{mol}^{-1}$ (ternary) for COGNAC, $0.024$ and $0.056\,\mathrm{mol}\,\mathrm{mol}^{-1}$ for HANNA, and $0.019$ and $0.025\,\mathrm{mol}\,\mathrm{mol}^{-1}$ for mod.~UNIFAC (Supplementary Note~\ref{SI-si:note_azeotrope_models}), whose parameters were fitted to the DDB azeotropic data~\cite{Gmehling1998}.

Fig.~\ref{fig:rcm_1} and Fig.~\ref{fig:rcm_2} show isothermal residue curve maps (RCMs) for two ternary mixtures. They trace the liquid composition during open evaporation and visualize the distillation regions defined by the pure components and azeotropes. COGNAC predicts all three binary azeotropes and the ternary azeotrope for acetone + chloroform + hexane (Fig.~\ref{fig:rcm_1}) with their measured types. 2-propanol + nitromethane + R112 (Fig.~\ref{fig:rcm_2}), outside the scope of mod.~UNIFAC, is split into three distillation regions by a ternary minimum-boiling azeotrope. The predicted ternary azeotropes deviate from the measured ones by at most $0.028$ and $0.018\,\mathrm{mol}\,\mathrm{mol}^{-1}$ in any mole fraction, and the predicted pressure deviates by $2.0\%$ and $1.8\%$, respectively.

\subsection{Model interpretability}
\label{sec:interpretability}
\begin{figure}[!h]
    \centering
        \includegraphics[width=\textwidth]{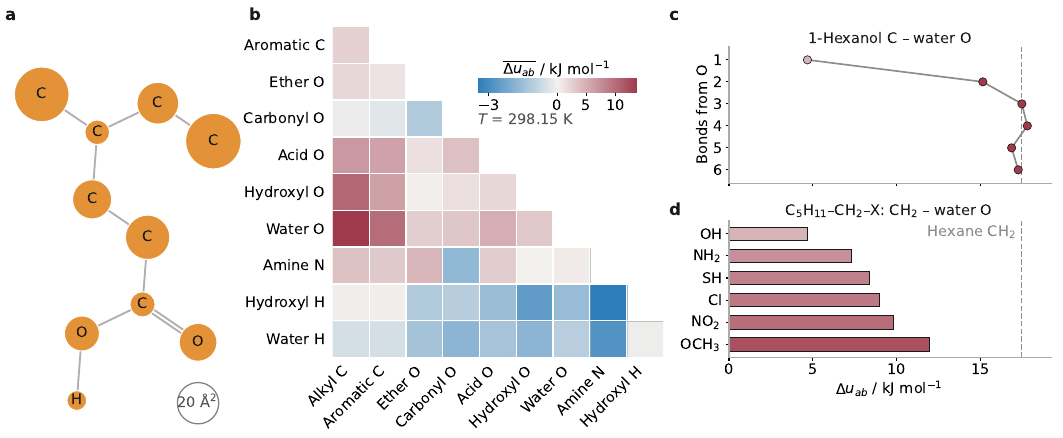}
        \begin{subfigure}[b]{0pt}\phantomsubcaption\label{fig:interpretability_geometry}\end{subfigure}%
        \begin{subfigure}[b]{0pt}\phantomsubcaption\label{fig:interpretability_interaction_matrix}\end{subfigure}%
        \begin{subfigure}[b]{0pt}\phantomsubcaption\label{fig:interpretability_neighborhood_chain}\end{subfigure}%
        \begin{subfigure}[b]{0pt}\phantomsubcaption\label{fig:interpretability_neighborhood_subst}\end{subfigure}%
    \caption{Interpretability of COGNAC. \textbf{a}, Molecular graph of 4-methylhexanoic acid. The surface area of the orange circles surrounding each atom is scaled with the predicted atom surface area from COGNAC; carbon-bound hydrogens are folded into their carbon atom, and the hollow gray circle has an area of 20\,\AA$^2$. \textbf{b}, Matrix of averaged exchange energies $\overline{\Delta u_{ab}}$ at 298.15 K between selected atom types, obtained from all atom pairs of all binary mixtures in the data set, including pairs of atoms of the same component. Pairs of the same atom type are not shown; their $\overline{\Delta u_{ab}}$ would not vanish either, since each type pools atoms with different molecular neighborhoods and $\Delta u_{ab} = 0$ holds only for two atoms with identical embeddings. The color scale is centered at zero and scaled separately to the most negative and the most positive value. \textbf{c}, Exchange energy $\Delta u_{ab}$ at 298.15\,K between each carbon atom of 1-hexanol and the oxygen atom of water, over the number of bonds between the carbon atom and the hydroxyl oxygen. \textbf{d}, Exchange energy between the CH$_2$ carbon atom bonded to the substituent X of 1-substituted hexanes (C$_5$H$_{11}$--CH$_2$--X) and the oxygen atom of water. The dashed lines in \textbf{c} and \textbf{d} mark the central CH$_2$ groups of hexane (C3 and C4, equivalent by symmetry); the colors follow the color scale of \textbf{b}.}
    \label{fig:interpretability}
\end{figure}

COGNAC learns interpretable atom-level volume and surface area parameters from the data, regularized towards COSMO cavity geometries (Methods). As an example, in Fig.~\ref{fig:interpretability_geometry}, the molecular graph of 4-methylhexanoic acid shows that an atom's predicted exposed surface area decreases with its number of covalent bonds, from the terminal CH$_3$ groups to the branching CH group and the carboxyl carbon. For 45,488 molecules outside our training and evaluation data sets, the predicted molecular volumes and surface areas deviate from the COSMO references by 11.0\,\AA$^3$ and 14.1\,\AA$^2$ on average (1.7\,\AA$^2$ per atom) (Extended Data Fig.~\ref{ed:fig_cosmo_geometry}).

The pair interactions in COGNAC are modeled through the symmetric exchange energy $\Delta u_{ab} = u_{ab}-0.5(u_{aa}+u_{bb})$, the difference between the unlike interaction a--b and the average of the two like interactions a--a and b--b (Methods); negative values mean that unlike contacts are favorable. For pairs of a hydrogen-bond donor and an acceptor atom, the sign of $\Delta u_{ab}$ is not entirely free: during training, a regularizer penalizes positive, i.e., repulsive, values for these pairs (Methods). We learn their magnitudes, as well as the sign and magnitude for all other atom pairs, solely from the thermodynamic data. Consistent with chemical intuition, the interaction between the water oxygen and the alkyl group (including the implicitly treated hydrogen atoms) has the highest value; the lowest values occur for the amine nitrogen with hydrogen atoms, reflecting the hydrogen bonds formed (Fig.~\ref{fig:interpretability_interaction_matrix}). Figs.~\ref{fig:interpretability_neighborhood_chain} and~\ref{fig:interpretability_neighborhood_subst} quantify this dependence on the molecular neighborhood for a single atom type. For the carbon atoms of 1-hexanol and the oxygen of water (Fig.~\ref{fig:interpretability_neighborhood_chain}), the exchange energy is 4.7\,kJ\,mol$^{-1}$ for the carbon bonded to the hydroxyl group and 15.1\,kJ\,mol$^{-1}$ for the next one; from the third carbon on, it is close to the value of an alkane carbon (17.4\,kJ\,mol$^{-1}$ for the central CH$_2$ groups of hexane). Thus, the hydroxyl group's influence extends over about two bonds. The CH$_2$ group next to other substituents also interacts more favorably with water oxygen than an alkane CH$_2$ group (Fig.~\ref{fig:interpretability_neighborhood_subst}). In mod.~UNIFAC, by contrast, all carbon atoms of 1-hexanol are CH$_2$ groups with identical interaction parameters, and the influence of a substituent on its neighboring CH$_2$ group is only captured where a larger predefined group, such as CH$_2$Cl or CH$_2$NH$_2$, includes it. 

\FloatBarrier
\subsection{Application in process simulation}
\label{sec:process_simulation}

The ultimate purpose of a predictive $g^\mathrm{E}$ model is to support process design and optimization, for which it must be integrated into process simulators. Here, COGNAC has a significant advantage over other ML-based models: Starting from only the SMILES of the components, COGNAC predicts the geometry parameters $R_a$ and $Q_a$ for every atom and the interaction parameters $a_{ab}, b_{ab}, c_{ab}$ for every pair. These parameters can then be used in the standard mod.~UNIFAC equation framework, which is implemented in all common process simulators and is cheap to evaluate. 

\begin{figure}[!h]
    \centering
    \includegraphics[width=\linewidth]{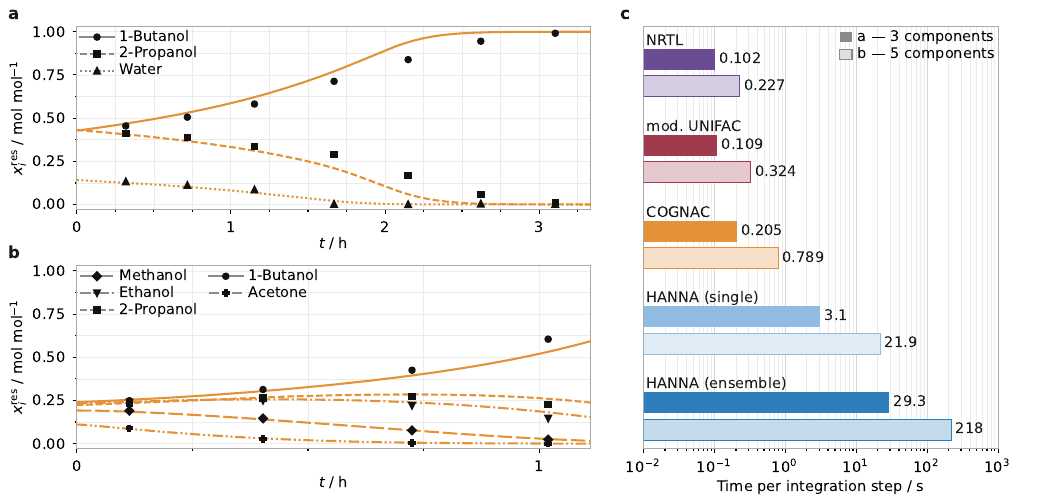}%
    \begin{subfigure}[b]{0pt}\phantomsubcaption\label{fig:batch_residue}\end{subfigure}%
    \begin{subfigure}[b]{0pt}\phantomsubcaption\label{fig:batch_residue_multi}\end{subfigure}%
    \begin{subfigure}[b]{0pt}\phantomsubcaption\label{fig:batch_runtime}\end{subfigure}%
    \caption{Dynamic simulation of a batch distillation using a Julia-based process simulator and the process model from Werner et al.~\cite{Werner2026}. \textbf{a,b}, Liquid mole fractions $x_i^\mathrm{res}$ in the reboiler, i.e., in the liquid residue, over the course of the batch distillation of a ternary mixture (1-butanol + 2-propanol + water, head pressure 70\,kPa; \textbf{a}) and a five-component mixture (methanol + ethanol + 2-propanol + 1-butanol + acetone, head pressure 50\,kPa; \textbf{b}), simulated with COGNAC (lines). Symbols denote experimental data from a laboratory-scale plant~\cite{Arweiler2026}. \textbf{c}, Wall-clock time that the integrator spends per time step for five thermodynamic models and the two mixtures. Median of up to 10 repeated solves; the compilation time of the generated residual functions is measured separately and excluded.}
    \label{fig:process_simulation}
\end{figure}

To demonstrate COGNAC's computational efficiency, we model batch distillations of a ternary and a five-component mixture in our laboratory-scale column~\cite{Arweiler2026} (Supplementary Note~\ref{SI-si:note_process_models}). The predictions agree well with the experimental data as shown in Figs.~\ref{fig:batch_residue} and \ref{fig:batch_residue_multi}. In Fig.~\ref{fig:batch_runtime}, we compare the time per integration step of the simulation using COGNAC with four other models for the same task. NRTL is the cheapest to evaluate, but it is not a predictive model and must be parameterized for every binary subsystem before the simulation. The computational costs of mod.~UNIFAC and COGNAC differ because they have different numbers of geometric and interaction parameters, a direct consequence of the different numbers of structural groups/atoms considered in the models. The chosen five-component mixture is an extreme example where mod.~UNIFAC heavily reuses the same groups in several components (e.g., CH$_3$ and CH$_2$ recur across the alcohols and the ketone), while COGNAC differentiates between almost all atoms. Nevertheless, the computational costs of using COGNAC and mod.~UNIFAC are of the same order of magnitude: per integration step, COGNAC needs 1.9 (ternary mixture) and 2.4 times (five-component mixture) the time of mod.~UNIFAC.
This demonstrates COGNAC's suitability for process simulations. We can further reduce computational cost by exploiting the symmetry of the COGNAC interaction parameters (e.g., $a_{ab} = a_{ba}$). 
However, we deliberately used the full symmetric matrix here to simulate its use in the standard mod.~UNIFAC framework without customizations. 

\FloatBarrier
\section{Discussion}

COGNAC shows that the accuracy gap between group-contribution methods like mod.~UNIFAC and modern machine learning models like HANNA is not a property of the local-composition equations but of the rigid, predefined groups. Replacing the groups with atoms whose parameters depend on their molecular neighborhood, learned end-to-end from 822,780 binary data points, makes the unchanged mod.~UNIFAC equations more accurate than HANNA for every data type of binary, ternary, and quaternary mixtures, including aqueous mixtures, and for azeotrope predictions, while keeping the simple deployment path of a group-contribution method. One network evaluation per mixture yields a parameter table valid for all temperatures and compositions, and a process simulation runs within a factor of about two of mod.~UNIFAC and two orders of magnitude faster than with the HANNA ensemble. Because the activity coefficients are derived from the excess Gibbs energy, they strictly satisfy the Gibbs--Duhem equation, and the mod.~UNIFAC equations ensure permutation symmetry and the correct pure-component limits. COGNAC robustly extends to any number of components without additional parameters or approximations.

The architecture also offers insight into the mixture physics with geometry parameters that were regularized against COSMO cavities and interaction parameters that align well with chemical intuition. This increases the interpretability over other ``black-box'' machine learning models, but also over the standard mod.~UNIFAC, where effects like hydrogen bonding are only implicitly contained in the group parameters.

COGNAC covers all molecules and ionic liquids built from the twelve elements in our data set (Methods). Mod.~UNIFAC describes only about half of the systems in our data; the other half, which contains many ionic liquids and multifunctional molecules, cannot be described with the published groups and parameters. Electrolytes and polymers are currently outside COGNAC's domain; however, extensions analogous to those of UNIFAC for long-range electrostatic interactions and free-volume effects are possible. The idea behind COGNAC generalizes beyond mod.~UNIFAC and applies to any group-contribution method.

We make the trained models and corresponding code available as open source (Code availability) and on our website \href{https://ml-prop.mv.rptu.de/}{MLPROP}, which provides an interactive user interface to calculate activity coefficients, excess enthalpies, and phase equilibria with COGNAC. Users can also download the parameter tables for any specified system for direct implementation in process simulators. The model is also available as a Model Context Protocol (MCP) connector for thermodynamic calculations in large language models.

\section{Methods}
\subsection{Equation framework}
\label{sec:equation_framework}
COGNAC is a hybrid model for excess Gibbs energy $g^\mathrm{E}$ that predicts the activity coefficient $\gamma_i$ in liquid mixtures as a function of temperature $T$ and composition $\mathbf{x}$. The activity coefficient $\gamma_i$ of component $i$ in the mixture is formally separated into a combinatorial part, which accounts for differences in molecular size and shape, and a residual part, which accounts for the interactions between molecular sites and contains the entire excess enthalpy:
\begin{equation}
\ln \gamma_i = \ln \gamma_i^{\mathrm{C}} + \ln \gamma_i^{\mathrm{R}}
\end{equation}
COGNAC computes both contributions from learned molecular representations shared across the two terms. For the combinatorial part, we adopt the expression of mod.~UNIFAC~\cite{Weidlich1987} with coordination number $z=10$:
\begin{equation}
\label{eq:comb_dortmund}
\ln \gamma_i^{\mathrm{C}}
= 1 - V_i' + \ln V_i'
- \frac{z}{2}\,q_i\left(1 - \frac{V_i}{F_i} + \ln\frac{V_i}{F_i}\right),
\end{equation}
with the volume and surface area fractions:
\begin{align}
V_i = \frac{r_i}{\sum_k x_k r_k},\qquad
V_i' = \frac{r_i^{3/4}}{\sum_k x_k r_k^{3/4}},\qquad
F_i = \frac{q_i}{\sum_k x_k q_k},
\end{align}
where $x_i$ is the mole fraction of component $i$ in the mixture. Here, $r_i$ and $q_i$ are the molecular volume and surface area parameters of component $i$, respectively. The exponent $3/4$ in $V_i'$ is the empirical correction to the Staverman--Guggenheim entropy of mixing that was introduced with mod.~UNIFAC. Setting $V_i' = V_i$ recovers the Staverman--Guggenheim expression of the original UNIFAC exactly:
\begin{equation}
\label{eq:comb_sg}
\ln \gamma_i^{\mathrm{C,SG}}
= \ln\!\left(\frac{\Phi_i}{x_i}\right)
+ \frac{z}{2}\,q_i\ln\!\left(\frac{\theta_i}{\Phi_i}\right)
+ \ell_i
- \frac{\Phi_i}{x_i}\sum_k x_k \ell_k,
\qquad \ell_i = \frac{z}{2}(r_i-q_i) - (r_i-1),
\end{equation}
with the volume fraction $\Phi_i = x_i V_i$ and the surface area fraction $\theta_i = x_i F_i$, so that the $3/4$ exponent is the only difference between the two combinatorial terms.

The molecular parameters $r_i$ and $q_i$ are obtained by summing over the contributions $R_a$ and $Q_a$ of all nodes of the molecular graph of component $i$ (all heavy atoms and explicit polar hydrogen atoms, see Section~\ref{sec:model_architecture}):
\begin{equation}
\label{eq:rq_sum}
r_i = \sum_{a\in\mathcal{V}_i} R_a,
\qquad
q_i = \sum_{a\in\mathcal{V}_i} Q_a.
\end{equation}
Here, $\mathcal{V}_i$ is the set of all nodes in the molecular graph of component $i$. Unlike UNIFAC, the $R_a$ and $Q_a$ values are not tabulated from predefined groups; instead, the model learns them end-to-end during training for each node, depending not only on atom type but also on its molecular neighborhood. As in, e.g., the UNIFAC groups CH$_3$, CH$_2$, and CH, whose parameters include bonded hydrogen atoms, the volume and surface area contributions of hydrogen atoms bonded to carbon are implicitly absorbed into the predicted $R_a$ and $Q_a$ of the parent carbon atom. By contrast, hydrogen atoms bonded to non-carbon atoms are represented as explicit nodes and receive their own volume and surface area parameters (see Section~\ref{sec:model_architecture}).

The residual contribution of component $i$ is computed as a sum of differences between the residual activity coefficients of its constituent atoms in the mixture and in the pure-component reference state:
\begin{equation}
\label{eq:residual}
\ln \gamma_i^{\mathrm{R}} =
\sum_{a\in\mathcal{V}_i}
Q_a\left(
\ln\Gamma_a^{\mathrm{mix}} - \ln\Gamma_a^{(i)}
\right),
\end{equation}
Here, $\Gamma_a$ is the residual activity coefficient of a single surface segment of atom $a$, evaluated in the mixture ($\Gamma_a^{\mathrm{mix}}$) and in the pure-component reference state of component $i$ ($\Gamma_a^{(i)}$). The weight $Q_a$ is the surface area of atom $a$ measured in units of the standard segment area. It thus counts the number of segments the atom contributes to the ensemble, so that every atom enters Eq.~\eqref{eq:residual} in proportion to the surface it exposes. $\mathcal{V}_i$ denotes the set of atomic graph nodes of component $i$, i.e., its heavy atoms and explicit polar hydrogen atoms (see Section~\ref{sec:model_architecture}). The segment residual activity coefficients $\ln\Gamma_a$ follow from the closed-form UNIFAC expression, applied to atoms instead of structural groups. It requires two quantities, both of which are outputs of the model parts described in the following sections: the surface area fraction $\Theta_a$ of every atom, which follows from the predicted surface area parameters $Q_a$ (Eq.~\eqref{eq:rq_sum}, Section~\ref{sec:volume_area_parameters}), and the Boltzmann factor $\Psi_{ab}$ of every atom pair, which follows from the predicted exchange energy $\Delta u_{ab}$ (Eq.~\eqref{eq:boltzmann_factor} below, Section~\ref{sec:interaction_parameters}). The surface area fractions in the mixture and in the pure-component reference state of component $i$ are calculated via
\begin{equation}
\label{eq:theta}
\Theta_a^{\mathrm{mix}} = \frac{Q_a\, x_{i(a)}}{\sum_j x_j\, q_j}, \quad
\Theta_a^{(i)} = \frac{Q_a}{q_i},
\end{equation}
respectively, where $i(a)$ denotes the component containing atom $a$. With the auxiliary sum
\begin{equation}
\label{eq:s_explicit}
s_b = \sum_{c\in\mathcal{V}} \Theta_c\,\Psi_{cb},
\end{equation}
the residual activity coefficient of a segment of atom $a$ is
\begin{equation}
\label{eq:ln_gamma_explicit}
\ln\Gamma_a
= 1 - \ln s_a - \sum_{b\in\mathcal{V}} \frac{\Theta_b\,\Psi_{ab}}{s_b}.
\end{equation}
Weighted by $Q_a$ in Eq.~\eqref{eq:residual}, this is the residual term of mod.~UNIFAC, in which the surface area parameter appears as a prefactor of the bracket. COGNAC therefore uses the equations of mod.~UNIFAC unchanged, with atoms taking the place of the structural groups. For $\Gamma_a^{\mathrm{mix}}$, the sums run over $\mathcal{V}=\bigcup_i\mathcal{V}_i$, i.e., over all atoms $b$ of all components present in the mixture, with $\Theta^{\mathrm{mix}}$, so that the same equations apply to a mixture with any number of components; for the pure-component reference $\Gamma_a^{(i)}$, they run over $\mathcal{V}_i$ only, with $\Theta^{(i)}$. All quantities on the right-hand side follow directly from $\Theta$ and $\Psi$ without iteration, so the residual term is cheap to evaluate and differentiable in a single forward pass. We also tested a self-consistent alternative to Eq.~\eqref{eq:ln_gamma_explicit}, the implicit closure of COSMOSPACE~\cite{Klamt2002}; we describe and evaluate this model variant, COGNAC-IMP (IMP: implicit closure), in Supplementary Note~\ref{SI-si:note_imp}, including the settings of its iterative solver (Supplementary Fig.~\ref{SI-si:fig_solver}) and its accuracy relative to COGNAC (Supplementary Table~\ref{SI-si:tab_imp_vs_exp}).

The Boltzmann factor $\Psi_{ab}$ is formed from the symmetric exchange energy $\Delta u_{ab}$ of the atom pair, i.e., the excess of the unlike $a$--$b$ interaction over the mean of the two like ($a$--$a$ and $b$--$b$) interactions, which is predicted by the model. Following mod.~UNIFAC, its temperature dependence is a quadratic polynomial:
\begin{equation}
\label{eq:boltzmann_factor}
\Psi_{ab} = \exp\!\left(-\frac{\Delta u_{ab}(T)}{R\,T}\right),
\qquad
\frac{\Delta u_{ab}(T)}{R} = A_{ab} + B_{ab}\,\tau + C_{ab}\,\tau^{2},
\qquad
\tau = \frac{T}{T_{\mathrm{ref}}} - 1 ,
\end{equation}
with $T_{\mathrm{ref}} = 298.15\ \mathrm{K}$. The three coefficients $A_{ab}$, $B_{ab}$, and $C_{ab}$ are temperature-independent, carry units of kelvin, and are predicted per atom pair by a permutation-invariant network from the two atom embeddings alone. They are the atom-pair analog of the tabulated group interaction parameters of mod.~UNIFAC. However, unlike the published mod.~UNIFAC parameters, they are derived using a symmetric exchange formulation, which ensures $\Delta u_{ab} = \Delta u_{ba}$ and $\Delta u_{aa} = 0$ and, hence, $\Psi_{aa} = 1$ exactly and at every temperature. Section~\ref{sec:interaction_parameters} gives a detailed derivation of the transformation between the two sets of coefficients and, with it, the route to a parameter table that a process simulator can use directly.

To summarize, the COGNAC equation framework requires the following quantities as input: (i) atom-level volume and surface area parameters $R_a$ and $Q_a$ for the combinatorial and residual contributions; (ii) atom-level pair exchange energies $\Delta u_{ab}(T)$ for the residual contribution. These quantities are predicted from the GNN-learned atom-level representations, as described in the following sections.

\subsection{Molecular graphs}
\label{sec:model_architecture}
Each component $i$ is represented as a molecular graph $\mathcal{G}_i=(\mathcal{V}_i,\mathcal{E}_i)$ constructed from a SMILES string using RDKit~\cite{rdkit}. The node set $\mathcal{V}_i$ comprises all heavy atoms of the molecule and, additionally, every hydrogen atom that is not bonded to a carbon atom, which is included as an explicit node; we refer to these as polar hydrogen atoms in the following. We treat only hydrogen atoms bonded to carbon implicitly; they are encoded only as a feature of the parent carbon atom. This hybrid treatment keeps the graphs compact for apolar molecules while explicitly resolving these polar hydrogens, which can, e.g., act as hydrogen-bond donors: this way, water, for example, is represented by three nodes instead of a single one, and alcohols, carboxylic acids, and amines expose their donor hydrogens as separate interaction sites. This lets the model describe hydrogen bonding through attractive pair interactions between distinct donor and acceptor sites rather than absorbing these effects into the parameters of the parent heavy atoms.

The edge set $\mathcal{E}_i$ contains all covalent bonds in component $i$, including bonds between explicit hydrogen nodes and their parent atoms, but excluding bonds between implicitly treated hydrogen atoms and their carbon atoms. For each node $a\in\mathcal{V}_i$, we compute six integer-valued categorical features: atomic number, total degree, formal charge, hybridization state, aromaticity, and the number of implicitly treated hydrogen atoms attached to the node. The total degree counts all bonded neighbors, including implicitly treated hydrogens. It is therefore not the degree of the node in the graph: a methyl carbon has total degree four but only one graph neighbor. Note that for heavy atoms whose hydrogens are explicit nodes (e.g., a hydroxyl oxygen), the last feature is always zero, as the hydrogens are instead resolved by the graph topology. For each covalent bond $a$--$b$, we encode three categorical edge features: bond type (single, double, triple, or aromatic), a binary conjugation flag, and a binary ring-membership flag. The hybridization state distinguishes sp, sp$^2$, sp$^3$, sp$^3$d, and sp$^3$d$^2$ and assigns all other states, in particular the s state of the explicit hydrogen nodes, to an additional category. The bond type includes an additional category for other bond types, which does not occur in the data set used here.

\subsection{Local message passing encoder}
\label{sec:encoder}
Each categorical atom and bond feature is mapped to a $D$-dimensional vector ($D=256$) via a dedicated learned embedding table. The per-feature embeddings are summed and layer-normalized to give the initial atom states $\mathbf{h}_a^{(0)}$ and bond states $\mathbf{e}_{ab}^{(0)}$:
\begin{equation}
\mathbf{h}_a^{(0)} = \mathrm{LN}\!\left(\sum_{f=1}^{6} \mathbf{E}_f\big[\chi_a^{(f)}\big]\right), \quad
\mathbf{e}_{ab}^{(0)} = \mathrm{LN}\!\left(\sum_{g=1}^{3} \mathbf{B}_g\big[\chi_{ab}^{(g)}\big]\right).
\end{equation}
Here, $\chi_a^{(f)}$ is the integer category of the $f$-th atom feature of atom $a$ and $\chi_{ab}^{(g)}$ that of the $g$-th bond feature of the bond $a$--$b$ (see Section~\ref{sec:model_architecture}). $\mathbf{E}_f$ and $\mathbf{B}_g$ are the learned embedding tables of these features, i.e., matrices with one $D$-dimensional row per category, and $\mathbf{E}_f[\chi]$ denotes the row of category $\chi$. $\mathrm{LN}$ denotes layer normalization~\cite{Ba2016}, which shifts and scales a vector to zero mean and unit variance over its $D$ entries and then applies a learned elementwise affine transformation.

We then refine the atom states with $L=4$ message-passing layers. At layer $\ell$, each atom $a$ aggregates messages from its covalently bonded neighbors $\mathcal{N}(a)$ and updates its state via a residual connection:
\begin{align}
\mathbf{m}_a^{(\ell)}
&=
\sum_{b\in\mathcal{N}(a)}
\phi_{\mathrm{msg}}^{(\ell)}
\!\left(
\mathbf{h}_a^{(\ell)},\mathbf{h}_b^{(\ell)},\mathbf{e}_{ab}^{(0)}
\right),
\label{eq:message}\\
\mathbf{h}_a^{(\ell+1)}
&=
\mathrm{LN}\!\left(
\mathbf{h}_a^{(\ell)} +
\phi_{\mathrm{upd}}^{(\ell)}\!\left(\mathbf{h}_a^{(\ell)},\mathbf{m}_a^{(\ell)}\right)
\right).
\label{eq:update}
\end{align}
Both $\phi_{\mathrm{msg}}^{(\ell)}$ and $\phi_{\mathrm{upd}}^{(\ell)}$ are two-layer multilayer perceptrons (MLPs) with sigmoid linear unit (SiLU) activations. The message (msg) function takes the concatenated central-atom state, neighbor state, and bond embedding as input ($3D \to D \to D$); the update (upd) function takes the concatenated atom state and aggregated message as input ($2D \to D \to D$). Bond embeddings $\mathbf{e}_{ab}^{(0)}$ are not updated across layers and serve as static edge context. After $L$ layers, $\mathbf{h}_a^{(L)} \in \mathbb{R}^D$ encodes the chemical identity of atom $a$ and its molecular neighborhood up to a radius of $L=4$ bonds. For brevity we use $\mathbf{h}_a \equiv \mathbf{h}_a^{(L)}$ below.

\subsection{Atom-level volume and surface area}
\label{sec:volume_area_parameters}

A shared two-output MLP head predicts two positive scalar quantities for each atom $a$ from its final embedding $\mathbf{h}_a$: an effective cavity volume $V_a$ and an effective cavity surface area $A_a$. Softplus activations ensure positivity:
\begin{align}
V_a &= \mathrm{softplus}\!\left(f_V(\mathbf{h}_a)\right) + \epsilon,\\
A_a &= \mathrm{softplus}\!\left(f_A(\mathbf{h}_a)\right) + \epsilon.
\end{align}
Here, $f_V$ and $f_A$ denote the two outputs of this head, an MLP with two hidden layers of widths $D$ and $D/2$ and SiLU activations, and $\epsilon = 10^{-8}$ (in \AA$^3$ and \AA$^2$, respectively) is a small constant that keeps both quantities strictly positive.

These atom-level quantities are normalized with global reference scales $V_{\mathrm{ref}}>0$ and $A_{\mathrm{ref}}>0$:
\begin{equation}
\label{eq:rq_norm}
R_a = \frac{V_a}{V_{\mathrm{ref}}},\quad
Q_a = \frac{A_a}{A_{\mathrm{ref}}} \quad \text{with} \quad
V_{\mathrm{ref}} = 20.18\ \text{\AA}^3,\quad A_{\mathrm{ref}} = 21.17\ \text{\AA}^2.
\end{equation}
We thereby follow the definition of the ``standard segment'' proposed by Abrams and Prausnitz for UNIQUAC~\cite{Abrams1975} and define a single CH$_2$ group in an infinitely long polyethylene molecule as the reference. However, unlike the original UNIQUAC formulation, our reference values are not the van der Waals volume and surface area of the standard segment, but the effective cavity volume and surface area from COSMO results in the CHAOS database~\cite{Gond2026}. During training, we also regularize the predicted geometries against the corresponding COSMO cavity values from the CHAOS database, which grounds the learned geometries in physical reality. We constrain the two quantities at different levels of resolution: the surface area is regularized per atom, whereas the volume is regularized per molecule; i.e., we compare the predicted total cavity volume of a component against its COSMO cavity volume. The per-atom COSMO area targets are constructed consistently with the graph representation: the cavity areas of implicitly treated (carbon-bonded) hydrogens are folded onto their parent carbon atom. In contrast, each explicit polar hydrogen retains its own cavity area. Section~\ref{sec:training} provides details on the regularization.

The molecular-level structural parameters $r_i$ and $q_i$ then follow by summing the atom-level contributions over the molecular graph, see Eq.~\eqref{eq:rq_sum}.

\subsection{Pairwise atom exchange energies}
\label{sec:interaction_parameters}
The three coefficients of the interaction polynomial in Eq.~\eqref{eq:boltzmann_factor} are predicted by the interaction head, a permutation-invariant neural network with three outputs (an MLP $3D \to D \to D/2 \to 3$ with SiLU activations) that takes as input the pairwise feature vector
\begin{equation}
\label{eq:interaction_features}
\mathbf{f}_{ab} = \left[
\mathbf{h}_a + \mathbf{h}_b,\quad
|\mathbf{h}_a - \mathbf{h}_b|,\quad
\mathbf{h}_a \odot \mathbf{h}_b
\right] \in \mathbb{R}^{3D},
\end{equation}
where $\odot$ denotes the elementwise product. All three components of $\mathbf{f}_{ab}$ are symmetric under permutation of $a$ and $b$, so the predicted coefficients, and hence $\Delta u_{ab}$, are symmetric. Note that temperature is not among the inputs: it enters only through the polynomial in Eq.~\eqref{eq:boltzmann_factor}, so the head predicts a temperature-independent parameter triple per atom pair.

In the first step, the interaction head outputs the three raw parameters $\tilde{A}_{ab}$, $\tilde{B}_{ab}$, and $\tilde{C}_{ab}$. Using the symmetric exchange formulation, these are transformed as follows:
\begin{equation}
\label{eq:exchange_coeffs}
X_{ab} = \tilde{X}_{ab} - \tfrac{1}{2}\left(\tilde{X}_{aa} + \tilde{X}_{bb}\right),
\qquad X \in \{A, B, C\},
\end{equation}
where the like-pair terms $\tilde{X}_{aa}$ and unlike-pair terms $\tilde{X}_{ab}$ are obtained from the same head. We apply the subtraction separately to each of the three channels. Because the mod.~UNIFAC polynomial is linear in the coefficients, this is equivalent to applying the exchange formulation on the interaction energy itself, with $u_{ab}(T)/R = \tilde{A}_{ab} + \tilde{B}_{ab}\,\tau + \tilde{C}_{ab}\,\tau^{2}$:
\begin{align} 
\label{eq:exchange_energy}\nonumber
\frac{\Delta u_{ab}(T)}{R} &= \frac{1}{R}\left[u_{ab}(T) - \tfrac{1}{2}\left(u_{aa}(T) + u_{bb}(T)\right)\right] \\ \nonumber
&= \left[\tilde{A}_{ab} + \tilde{B}_{ab}\,\tau + \tilde{C}_{ab}\,\tau^{2}\right] - \frac{1}{2}\left[\tilde{A}_{aa} + \tilde{B}_{aa}\,\tau + \tilde{C}_{aa}\,\tau^{2} + \tilde{A}_{bb} + \tilde{B}_{bb}\,\tau + \tilde{C}_{bb}\,\tau^{2}\right] \\ \nonumber
& = \tilde{A}_{ab} - \tfrac{1}{2}\left(\tilde{A}_{aa} + \tilde{A}_{bb}\right) + \left[\tilde{B}_{ab} - \tfrac{1}{2}\left(\tilde{B}_{aa} + \tilde{B}_{bb}\right)\right]\tau + \left[\tilde{C}_{ab} - \tfrac{1}{2}\left(\tilde{C}_{aa} + \tilde{C}_{bb}\right)\right]\tau^2 \\
&= A_{ab} + B_{ab}\tau + C_{ab}\tau^2
\end{align}

$\Delta u_{ab}$ is thus an exchange energy as defined in COSMOSPACE~\cite{Klamt2002}, with its temperature dependence constrained to the quadratic form of mod.~UNIFAC. The coefficients of Eq.~\eqref{eq:exchange_coeffs} are converted from the reduced basis $(1,\tau,\tau^{2})$ used internally to the published basis $(1,T,T^{2})$ by the exact change of basis
\begin{equation}
\label{eq:basis_change}
a_{ab} = A_{ab} - B_{ab} + C_{ab},\qquad
b_{ab} = \frac{B_{ab} - 2C_{ab}}{T_{\mathrm{ref}}},\qquad
c_{ab} = \frac{C_{ab}}{T_{\mathrm{ref}}^{2}},
\end{equation}
where $(a_{ab}, b_{ab}, c_{ab})$ are the coefficients for a mod.~UNIFAC group interaction table in which every heavy atom and every explicit hydrogen atom of the mixture is a group. Because the exchange subtraction is performed before conversion, the exported coefficients always satisfy $a_{aa} = b_{aa} = c_{aa} = 0$. Note that, e.g., the coefficients between two CH2 groups are not necessarily zero if their molecular neighborhoods are different. The obtained interaction coefficient matrix is always symmetric, i.e., $a_{ab} = a_{ba}$, $b_{ab} = b_{ba}$, and $c_{ab} = c_{ba}$.

The reduced basis in Eq.~\eqref{eq:boltzmann_factor} is necessary for stable training, since over the data's temperature range $\tau$ stays of order one whereas $T$ and $T^{2}$ are of order $10^{2}$ and $10^{5}$, respectively. As a single network emits all three coefficients, the original basis makes the three output channels differ in gradient magnitude by up to five orders of magnitude, and the resulting $\Delta u_{ab}(T)$ is already far outside the physical range at initialization; in the reduced basis, the three channels share one scale. The output weights of the $B$ and $C$ channels are initialized to zero, so the model starts with a temperature-independent interaction parameter and develops temperature dependence only where the data require it. This mirrors the published table of mod.~UNIFAC, where most group pairs have $c=0$, and many have $b=0$.

While the message-passing encoder sees each component's molecular graph separately, the interaction head operates on the full mixture, which includes all atomic nodes from all components. Thus, every pair of atoms in the mixture is evaluated, irrespective of whether the two atoms belong to the same component or to different ones, and including the self-pairs $a = b$, which supply the like-pair terms of Eq.~\eqref{eq:exchange_coeffs}. This is required because the interaction between all atoms in the mixture must be calculated to obtain the residual activity coefficients, following the equation framework described in Section~\ref{sec:equation_framework}. We evaluate the interaction head a second time on each isolated component, which supplies the $\Psi_{ab}$ for $\Gamma_a^{(i)}$ in the pure-component reference state.

Because donor hydrogens are explicit nodes, the model resolves a hydrogen bond between a donor hydrogen and an acceptor atom as an explicit pair interaction. In our experiments, we noticed that the model did not always assign this interaction the expected strongly attractive (i.e., negative) value for $\Delta u_{ab}$, but, due to its flexibility, it occasionally spread this interaction onto other attached atoms. For this reason, we introduced a regularization term that encourages the correct sign of hydrogen-bond interactions (see Section~\ref{sec:training}). Aside from this, we determine all learned interactions from experimental data alone.

\subsection{Data processing and splitting}
\label{sec:data}

We obtained all experimental data used to train and evaluate the model from the Dortmund Data Bank (DDB) \cite{DDB2026}. Vapor--liquid equilibrium data were adopted from the ``VLE'' and ``HPV'' (high-pressure equilibria) databases. We also used the following databases: ``ACT'' (activity coefficients at infinite dilution), ``LLE'' (liquid--liquid equilibria), ``HE'' (excess enthalpies), and ``SLE'' (solid--liquid equilibria). We used Antoine parameters to calculate pure-component vapor pressures (supplemented by measured pure-component points where the VLE data sets report them). We also obtained melting temperatures (``PCP\_MPT'') and melting enthalpies (``PCP\_HFS'') from the DDB. 

In the first step of data preprocessing, we unified all extracted data by column name, converted to SI units as needed, and reordered components so that all data for the same system used the same component order. One exception is the ACT data, where component order matters and the second component was always defined as the one at infinite dilution. For every molecule in the data set, we generated a canonical SMILES using RDKit~\cite{rdkit}. In this process, we excluded all rows containing a component without a valid SMILES. The data set covers molecular components and ionic liquids; all molecules are built from the elements H, B, C, N, O, F, Si, P, S, Cl, Br, and I. In the following, the type-specific processing of the raw data is described for binary mixtures, which form the entire training data. We used the data for ternary and quaternary mixtures only for evaluation and converted them in the same way, extending them to the additional components. We carried out the manual inspection described below only for binary data.

\paragraph{VLE and HPV.} In the VLE and HPV data, only points at finite dilution were considered, and the pure-component edges ($x_1 \in \{0,1\}$ or $y_1 \in \{0,1\}$) were removed. Where an isothermal data set itself reports a measured pure-component point ($x_1 \in \{0,1\}$), we used its total pressure directly as the pure-component vapor pressure $p_i^{\mathrm{s}}$. Otherwise, we computed $p_i^{\mathrm{s}}$ from the DDB Antoine equation

\begin{equation}
  \log_{10}\frac{p_i^{\mathrm{s}}}{\mathrm{mmHg}} = A_i - \frac{B_i}{C_i + \vartheta/{}^\circ\mathrm{C}},
\end{equation}

using only parameter sets whose temperature range contains $\vartheta$. For isobaric data sets that report a pure-component point, we recalibrated the Antoine equation to it: we divided the measured pressure at this point by the Antoine vapor pressure at its measured temperature (i.e., the pure-component boiling temperature) and multiplied $p_i^{\mathrm{s}}$ of all data points of the data set by this factor. This is equivalent to shifting $A_i$ such that the Antoine equation passes through the measured point while retaining its temperature dependence. Using the edge points and recalibration substantially improved the quality of the derived activity coefficients, especially near the pure-component limit, where they must converge to unity. We dropped data points for which $p_i^{\mathrm{s}}$ could be obtained in neither way.

The activity coefficients were derived using extended Raoult's law:
\begin{equation}
  \gamma_i = \frac{p\, y_i}{p_i^{\mathrm{s}}\,x_i}
  \label{eq:raoult}
\end{equation}
Because this assumes an ideal vapor phase, we excluded all data points with total pressure above 10 bar.

Rows are then classified into two data types for the training: First, a row with $T$, $p$, $x_1$, $y_1$, and both vapor pressures $p^s_1(T)$, $p^s_2(T)$ at the reported temperature yields both activity coefficients and is labeled as TPXY. Second, we label a row missing vapor-phase composition information as TPX and retain it without activity coefficients, since it still contains the total pressure as a valid training target. All other rows (e.g., without pressure information) were dropped.

\paragraph{LLE.} For the LLE data, the DDB reports the temperature $T$, the pressure $p$ (if no value was reported, we assumed atmospheric pressure), and the compositions of the liquid phases in equilibrium, $x_1'$ and $x_1''$. If the composition of both phases is reported in the DDB, we processed the data such that $x_1' <x_1''$ always holds. If only one phase composition was reported, we kept the DDB phase assignment, as it was correct in almost all cases. Because the COGNAC framework does not include pressure dependence, we excluded data points with pressures above 50 bar.

\paragraph{SLE.} We used the following equation to calculate activity coefficients from the SLE data:
\begin{equation}
  \gamma_i = \frac{1}{x_i}\,
  \exp\!\left[\frac{\Delta h_{\mathrm{fus},i}}{R\,T_{\mathrm{fus},i}}\,
  \frac{T - T_{\mathrm{fus},i}}{T}\right].
  \label{eq:sle}
\end{equation}
We thereby used the average of all reported melting temperatures $T_{\mathrm{fus},i}$ and enthalpies of fusion $\Delta h_{\mathrm{fus},i}$ for pure component $i$ in the DDB (below $50\,\mathrm{bar}$ or with no reported pressure). This formula holds only on a liquidus branch where the pure component $i$ is in equilibrium with the liquid phase. To identify the branches, we manually checked all reported mixtures with at least 5 data points. We saved the composition ranges of each eligible liquidus branch (i.e., the range for which the liquid is actually in equilibrium with a pure component solid phase). We used only data from these branches to calculate SLE activity coefficients; all other data were dropped.

\paragraph{Manual inspection.} Before training, we visually inspected the binary data mixture by mixture, with simple fits to the data of each mixture as a guide. We rejected individual outliers as well as entire data sets that were internally inconsistent or contradicted the other data of the mixture, and accepted the remaining data points. For VLE, HE, and LLE, mixtures whose data are near-ideal or closely described by such a fit were accepted automatically. Only mixtures with enough data points for such a comparison were inspected (at least 5 for SLE, HE, and LLE, and at least 10 for VLE and ACT); the data points of all other mixtures are labeled ``unreviewed'' and retained. For SLE, the inspection also identified the composition range of each liquidus branch, and only these branches are used. No model predictions were used. Details are given in Supplementary Note~\ref{SI-si:note_data}.

\paragraph{Data split.} The pool of data for binary mixtures used for training and evaluating COGNAC contains all accepted and unreviewed data points (i.e., all data points except those rejected in the review); in total, 822,780 data points for 48,432 binary systems were used. To assess COGNAC's prediction performance on the binary data, we split this data set system-wise into ten folds for ten-fold cross-validation. Folds were assigned per pair of canonical SMILES such that the same binary systems never fall into different folds. The assignment was stratified such that every data type is distributed evenly across the folds (greedy assignment, random seed 0). In run $k$ of the cross-validation, fold $k$ is used as the test set, fold $k+1$ (modulo ten) as the validation set, and the remaining eight folds for training. Every system is thus predicted exactly once by a model that has not seen it, and all binary results in this work are these out-of-fold predictions. Supplementary Tables~\ref{SI-si:tab_data_split_points} and~\ref{SI-si:tab_data_split_systems} summarize the number of data points and systems per fold and data type. All binary results are evaluated on the accepted data points only, except for the ACT data, for which most systems contain too few data points for a review, and accepted and unreviewed data points are both used; results including the unreviewed data points of the other data types are given in Supplementary Note~\ref{SI-si:note_data} (Supplementary Fig.~\ref{SI-si:fig_review_status} and Supplementary Table~\ref{SI-si:tab_review_lle}). 

The ternary and quaternary data sets are evaluated on all data points from the DDB, except for 18 ternary LLE data points rejected in an inspection of the ternary LLE data. The quaternary data set comprises 21,221 TPXY, TPX, and LLE data points of 1,032 systems and is evaluated in Extended Data Fig.~\ref{ed:fig_quaternary}. The ACT data of the ternary data set (solute at infinite dilution in a mixed solvent) also include 1,674 data points in which the solvent is a pure component; since these are binary activity coefficients at infinite dilution, they were excluded from all ternary results. For the evaluation of ternary and quaternary mixtures, the prediction of ternary azeotropes, the interpretability analysis, and the process simulations, COGNAC was retrained on the full binary data (full-data model), using nine folds (739,675 data points) for training and the tenth fold (83,105 data points) as the validation set, without a test set, since none of these evaluations use binary test data. 

Both ablation studies follow the protocol above: the architectural ablation (Supplementary Note~\ref{SI-si:note_ablation}) uses ten-fold cross-validation on the binary mixtures, and the hydrogen ablation (Extended Data Fig.~\ref{ed:hydrogen_ablation}) additionally uses the full-data model for the ternary mixtures.

\subsection{Model training}
\label{sec:training}

COGNAC is trained jointly on six types of thermodynamic data for binary mixtures, which differ in what observable is directly measured and how the model prediction enters the loss. In all cases, the Huber (smooth L1) loss function is used:
\begin{equation}
\label{eq:huber}
\ell_\delta(r) =
\begin{cases}
\dfrac{r^2}{2\delta} & |r| < \delta, \\[4pt]
|r| - \dfrac{\delta}{2} & |r| \geq \delta,
\end{cases}
\end{equation}
with $r$ being the loss quantity and the threshold $\delta = 1$ unless stated otherwise. The Huber loss matches the curvature of mean squared error near the origin while degrading to mean absolute error for large residuals, reducing sensitivity to outliers in experimental data. Throughout this section, a hat denotes a thermodynamic property predicted by the model (e.g., $\ln\widehat\gamma$) to distinguish it from the corresponding experimental value. Internal model parameters, such as the cavity geometries $A_a$ and $r_i$ or the interaction energies $\Delta u_{ab}$, have no experimental counterpart and are therefore written without a hat.

Let $\mathcal{B}_D$ denote the set of data points of type $D$ in a mini-batch, and let $b\in\mathcal{B}_D$ be one such data point. The contribution of data type $D$ is the mean of a data-point-wise loss $\ell^D_b$ over $\mathcal{B}_D$,
\begin{equation}
\label{eq:batch_reduction}
\mathcal{L}_D = \frac{1}{|\mathcal{B}_D|}\sum_{b\in\mathcal{B}_D} \ell^D_b ,
\end{equation}
which defines the terms $\mathcal{L}_{\mathrm{TPXY}}$, $\mathcal{L}_{\mathrm{SLE}}$, $\mathcal{L}_{\mathrm{ACT}}$, $\mathcal{L}_{\mathrm{TPX}}$, $\mathcal{L}_{\mathrm{HE}}$, and $\mathcal{L}_{\mathrm{LLE}}$ of the total training objective, Eq.~\eqref{eq:total_loss}.
Targets are missing in parts of the data: an SLE data point yields only the activity coefficient of the component that is in equilibrium with its pure solid, and an LLE data point often reports only one branch of the tie line. Wherever a data point carries several targets, $\ell^D_b$ is therefore the mean of the expression given below over the finite targets of that data point. Since Eq.~\eqref{eq:batch_reduction} performs the same reduction for every data type, we state only the per-target expressions below.

\paragraph{TPXY and SLE.}
Both data types provide activity coefficients at a given composition and temperature, which the model predicts directly, so both enter through the same residual
\begin{equation}
\label{eq:loss_gamma}
\ell_1\!\left({\ln\widehat\gamma}_{b,i} - \ln\gamma_{b,i}\right),
\qquad i \in \{1,2\}.
\end{equation}
For the TPXY data, both $\ln\gamma_1$ and $\ln\gamma_2$ are obtained from the VLE measurements through the extended Raoult's law, Eq.~\eqref{eq:raoult} (Section~\ref{sec:data}). The SLE data give, via Eq.~\eqref{eq:sle}, the activity coefficient of the component in equilibrium with its pure solid, i.e., in general only for one of the two components. Both terms enter the total objective unweighted and together anchor the scale of the predicted activity coefficients. For the TPXY term, the mean in Eq.~\eqref{eq:batch_reduction} is taken directly over all activity coefficients in the mini-batch, so that each activity coefficient is weighted equally.

\paragraph{ACT.}
Each ACT data point directly provides the activity coefficient at infinite dilution, $\ln\gamma_2^{\infty}$, of a solute (component~2) in a solvent (component~1). For training on these data, the model is evaluated at the composition $x_1 = 1$, and only the predicted activity coefficient of the solute enters the loss,
\begin{equation}
\label{eq:loss_act}
\ell_1\!\left(\ln\widehat\gamma_{b,2}^{\infty} - \ln\gamma_{b,2}^{\infty}\right).
\end{equation}

\paragraph{TPX.}
TPX data are bubble-line measurements that report the total pressure $p$ at a given liquid composition $x_1$ and temperature $T$, but not the vapor composition, so they cannot be used to derive activity coefficients. Assuming an ideal vapor phase, the total pressure predicted from the predicted activity coefficients is
\begin{equation}
\label{eq:tpx_pressure}
\hat{p} = p_1^{\mathrm{s}}(T)\,x_{1,b}\,\hat\gamma_{1,b} + p_2^{\mathrm{s}}(T)\,x_{2,b}\,\hat\gamma_{2,b},
\end{equation}
and the residual is taken on its logarithm, so that data points spanning several orders of magnitude in pressure contribute comparably,
\begin{equation}
\label{eq:loss_tpx}
\ell_1\!\left(\ln(\hat{p}/\mathrm{bar}) - \ln(p/\mathrm{bar}) \right).
\end{equation}

\paragraph{HE.}
Excess enthalpy data, $h^\mathrm{E}$ in kJ\,mol$^{-1}$, relate to $g^\mathrm{E}$ through the Gibbs--Helmholtz equation
\begin{equation}
\label{eq:gibbs_helmholtz}
h^\mathrm{E} = -RT^2\!\left(\frac{\partial\,(g^\mathrm{E}/RT)}{\partial T}\right)_{p,x}.
\end{equation}
Since $g^\mathrm{E}/RT = \sum_i x_i \ln\gamma_i$ and the model predicts $\ln\gamma_i$ as a differentiable function of $T$, the derivative in Eq.~\eqref{eq:gibbs_helmholtz} is evaluated by automatic differentiation with respect to the input temperature,
\begin{equation}
\label{eq:he_pred}
\hat{h}^E_b = -R\,T_b^2\,\frac{\partial}{\partial T}\!\left(\sum_i x_{i,b}\,{\ln\widehat\gamma}_{i,b}\right),
\end{equation}
and the residual is
\begin{equation}
\label{eq:loss_he}
\ell_1\!\left(\hat{h}^E_b - h^\mathrm{E}_b\right).
\end{equation}
This term back-propagates gradients through the temperature derivative of the model output, so it trains temperature sensitivity directly, rather than only through the temperature dependence of the activity coefficients.

\paragraph{LLE.}
Liquid--liquid equilibrium data provide the compositions of the two coexisting liquid phases, $x_1'$ and $x_1''$. The model does not predict these compositions directly; they follow from the phase equilibrium conditions applied to the predicted liquid-phase Gibbs energy of mixing,
\begin{equation}
\label{eq:dgmix}
\frac{\Delta g_{\mathrm{mix}}}{RT} = \sum_i x_i\left(\ln x_i + \ln\gamma_i(\mathbf{x},T)\right),
\end{equation}
and must therefore be obtained from it by a solver that is itself differentiable. We use the convex envelope method (CEM)~\cite{Ryll2012} in a first step to obtain an approximate but non-differentiable solution. We evaluate the $\Delta g_{\mathrm{mix}}$ graph on a uniform grid of $G = 401$ compositions in one batched model call. The phase split is read off the lower convex hull of the resulting curve: an interval over which the hull lies strictly below $\Delta g_{\mathrm{mix}}$ is a miscibility gap. Its two endpoints are the binodal compositions. Two separate miscibility gaps at the same temperature appear as two separate intervals, since the hull touches $\Delta g_{\mathrm{mix}}$ between them; in this rare case, we compare the widest gap with the data point. Subsequently, we use a single Newton step to obtain a differentiable solution. For a data point $b$, we write the two unknowns of the binary phase split as $\mathbf{w}_b = (x_{1,b}',\,x_{1,b}'')$, with the corresponding phase compositions $\mathbf{x}_b' = (x_{1,b}',\,1-x_{1,b}')$ and $\mathbf{x}_b'' = (x_{1,b}'',\,1-x_{1,b}'')$. The phase split predicted by the model with parameters $\theta$ is a root of the isoactivity residual
\begin{equation}
\label{eq:isoactivity}
F_i(\mathbf{w}_b;\theta) = x_{i,b}'\,\gamma_i(\mathbf{x}_b',T_b;\theta) - x_{i,b}''\,\gamma_i(\mathbf{x}_b'',T_b;\theta), \qquad i = 1,2,
\end{equation}
and the grid result $\mathbf{w}_{\mathrm{CEM},b}$ is refined by one Newton step on this residual,
\begin{equation}
\label{eq:cem_newton}
\hat{\mathbf{w}}_b = \mathbf{w}_{\mathrm{CEM},b} - \mathbf{J}_b^{-1}\,F\!\left(\mathbf{w}_{\mathrm{CEM},b};\theta\right),
\qquad
\left(\mathbf{J}_b\right)_{ij} = \left.\frac{\partial F_i}{\partial w_j}\right|_{\mathbf{w}_{\mathrm{CEM},b}},
\end{equation}
in which $\mathbf{w}_{\mathrm{CEM},b}$ and $\mathbf{J}_b$ are treated as constants and only $F$ retains its dependence on $\theta$. The result $\hat{\mathbf{w}}_b = (\hat{x}'_{1,b},\,\hat{x}''_{1,b})$ is the predicted phase split. 

The Newton step serves two purposes. First, the grid result is only accurate to the grid spacing and, being a grid point, does not change with $\theta$, so that it provides no gradient; the Newton step moves the prediction towards the root of $F$, so that the loss is evaluated at the composition to which its gradient refers. Second, differentiating Eq.~\eqref{eq:cem_newton} by ordinary backpropagation yields $\mathrm{d}\hat{\mathbf{w}}_b/\mathrm{d}\theta = -\mathbf{J}_b^{-1}\,\partial F/\partial\theta$, which is the implicit function theorem gradient $-\left(\partial F/\partial\mathbf{w}\right)^{-1}\partial F/\partial\theta$ of the phase split, with the Jacobian evaluated at the grid result. It equals the exact implicit derivative only when the grid result coincides with the root; its accuracy therefore depends on the grid error and on the conditioning of $\mathbf{J}$. A small Tikhonov term ($10^{-4}\,\mathbf{I}$) is added to $\mathbf{J}$ before it is inverted, the step is restricted to a trust region of $0.05$ in mole fraction, and data points whose predicted gap is narrower than $0.01$ are left at their grid value, since $\mathbf{J}$ becomes singular as the critical point is approached. Because the condition number of $\mathbf{J}$ reaches $\mathcal{O}(10^4)$ for strongly associating systems, we evaluate and differentiate this part of the computational graph in full single precision, even when reduced-precision matrix multiplication is enabled elsewhere.

The residual is the difference between the two predicted phase compositions and the measured ones. Since a deviation in mole fraction can never exceed one, the default threshold $\delta = 1$ would keep the Huber loss quadratic over the entire possible range; we therefore use $\delta = 0.35$, so that large deviations enter linearly,
\begin{equation}
\label{eq:loss_lle}
\ell_{0.35}\!\left(\hat{x}'_{1,b} - x'_{1,b}\right)
\quad\text{and}\quad
\ell_{0.35}\!\left(\hat{x}''_{1,b} - x''_{1,b}\right).
\end{equation}
If the hull finds no miscibility gap, the model predicts a homogeneous mixture, no phase compositions exist, and the data point contributes zero. This has an important consequence: $\mathcal{L}_{\mathrm{LLE}}$ can refine a phase split that the model already predicts, but it can never create one: the detection of the gap from the convex hull is a discrete operation that is not differentiable, so the grid in the first step is evaluated without tracking gradients and no gradient reaches $\theta$ through the detection of the gap itself. At the beginning of training, the predicted $\Delta g_{\mathrm{mix}}$ is close to ideal and hence convex everywhere; no data point splits, and $\mathcal{L}_{\mathrm{LLE}}$ contributes no gradient at all, so that liquid--liquid demixing is initially learned only indirectly, through the other data types.

In our previous work~\cite{Hoffmann2026}, we addressed this problem by introducing a ``Gibbs loss'' that encourages the model to produce a correctly shaped $\Delta g_\mathrm{mix}$ by enforcing a negative second derivative on LLE data points during training. This idea was further developed by Ben Hicham et al.~\cite{Hicham2026}, who constrain the local curvature of $\Delta g_{\mathrm{mix}}$ at the compositions at which a phase split was measured: both measured binodal compositions should lie in a locally stable region of the surface and the midpoint of the gap, $z_b = \tfrac{1}{2}(x'_{1,b}+x''_{1,b})$, in an unstable one. While this condition is neither necessary nor sufficient for a phase split~\cite{Hicham2026}, the penalty based on it provides a gradient towards demixing where the model does not yet predict a phase split. For a binary mixture the composition Hessian of $\Delta g_{\mathrm{mix}}$ has a single independent element,
\begin{equation}
\label{eq:hessian_curvature}
H_b(x_1) = \frac{\partial^2}{\partial x_1^2}\!\left(\frac{\Delta g_{\mathrm{mix}}}{RT}\right)_b ,
\end{equation}
and the data-point-wise loss is the sum of three hinge penalties,
\begin{equation}
\label{eq:loss_hessian}
\ell^{\mathrm{H}}_b =
\mathrm{ReLU}\!\left(\varepsilon_{\mathrm{H}} - H_b(x'_{1,b})\right)
+ \mathrm{ReLU}\!\left(\varepsilon_{\mathrm{H}} - H_b(x''_{1,b})\right)
+ \mathrm{ReLU}\!\left(H_b(z_b) + \varepsilon_{\mathrm{H}}\right),
\end{equation}
where $\varepsilon_{\mathrm{H}} = 0.1$ is the margin each of the three inequalities must exceed before the corresponding penalty vanishes. $H_b$ is computed exactly by twofold automatic differentiation of the predicted $\Delta g_{\mathrm{mix}}$ with respect to composition, so that Eq.~\eqref{eq:loss_hessian} remains differentiable with respect to $\theta$.

This term is the counterpart of $\mathcal{L}_{\mathrm{LLE}}$, not a duplicate. Because it is evaluated at measured compositions, it has a gradient on precisely those data points for which the solver found no split and $\mathcal{L}_{\mathrm{LLE}}$ is identically zero, including at initialization. It is therefore applied to every LLE data point during training, not only to those the solver failed to split. The interplay of the two terms is directly visible in the training history: the logged $\mathcal{L}_{\mathrm{LLE}}$ \emph{rises} over the first few epochs before it falls, because data points that previously predicted no split begin to split and enter the loss for the first time. 

Much of the LLE data is single-branch mutual-solubility data for which only one endpoint is reported, so $z_b$ is undefined. Such a data point contributes only the hinge penalty at the endpoint that was measured. Since an ideal mixture already satisfies this penalty, it does not encourage demixing by itself; single-branch data points shape the predicted gap only through $\mathcal{L}_\mathrm{LLE}$ once the model predicts a split.

\paragraph{Geometry regularization.}
We regularize the predicted cavity geometries against the COSMO reference values in the CHAOS database~\cite{Gond2026} (see Section~\ref{sec:volume_area_parameters}) during training. The penalty is taken in the logarithm, so that it constrains the relative rather than the absolute deviation and hence does not concentrate on the largest atoms and molecules,
\begin{equation}
\label{eq:loss_geom}
\mathcal{L}_{\mathrm{geom}} =
\left\langle\left(\ln A_a - \ln A_a^{\mathrm{COSMO}}\right)^2\right\rangle_a
+
\left\langle\left(\ln\!\left(r_i\,V_{\mathrm{ref}}\right) - \ln V_i^{\mathrm{COSMO}}\right)^2\right\rangle_i ,
\end{equation}
where $\langle\cdot\rangle_a$ averages over the atoms and $\langle\cdot\rangle_i$ over the components of the mini-batch for which a reference value is available. As discussed in Section~\ref{sec:volume_area_parameters}, the surface area is constrained per atom whereas the cavity volume is constrained per molecule, i.e., on the summed molecular volume $r_i V_{\mathrm{ref}}$. COSMO results are available only for a subset of molecules in the training set; atoms and molecules without a reference value are masked out and contribute nothing to Eq.~\eqref{eq:loss_geom}, so the loss aligns predicted geometries with COSMO cavities on the remaining molecules as a generalization rather than a fit.

\paragraph{Hydrogen-bond sign regularizer.}
Under the explicit polar-hydrogen graph representation (Section~\ref{sec:model_architecture}), a hydrogen bond between a donor and an acceptor atom is resolved as an individual atom-pair interaction, which should therefore be attractive, i.e., $\Delta u_{ab} < 0$ (Eq.~\eqref{eq:boltzmann_factor}). Without an explicit physical prior, we found that the model instead distributed the effect of hydrogen bonding across several neighboring atoms, yielding inconsistent, sometimes repulsive, individual pair energies, so the sign of $\Delta u_{ab}$ for an individual donor/acceptor pair carried no reliable physical meaning. We therefore add a hinge penalty that discourages repulsive interactions between hydrogen-bond donor and acceptor atoms. We consider an atom as a donor if it is an explicit hydrogen node bonded to N, O, or S; explicit hydrogen nodes bonded to any other atom are not counted as donors. Furthermore, we consider an atom as an acceptor if it is oxygen, or a nitrogen without positive formal charge that is neither a pyrrole-type aromatic N--H (whose lone pair is delocalized into the aromatic ring) nor part of a formally charged aromatic ring (e.g., imidazolium, pyridinium), in which the charge is delocalized onto the ring nitrogen. For every donor/acceptor atom pair $(a,b)$ found in a mixture batch, the penalty is
\begin{equation}
\label{eq:hbond_loss}
\mathcal{L}_{\mathrm{hbond}} = \frac{1}{|\mathcal{P}|}\sum_{(a,b)\in\mathcal{P}} \left[\mathrm{ReLU}\!\left(\frac{\Delta u_{ab}}{\Delta u_{\mathrm{scale}}}\right)\right]^2,
\qquad \Delta u_{\mathrm{scale}} = R \cdot 100\ \mathrm{K} \approx 0.83\ \mathrm{kJ}\,\mathrm{mol}^{-1},
\end{equation}
where $\mathcal{P}$ is the set of donor/acceptor atom pairs in the batch (each unordered pair counted once), so the penalty is only nonzero when a donor/acceptor pair is repulsive ($\Delta u_{ab} > 0$); the constant $\Delta u_{\mathrm{scale}}$ only sets the scale of the term relative to the others, as $\Delta u_{ab}$ reaches the order of $10\ \mathrm{kJ}\,\mathrm{mol}^{-1}$. This term is included with weight $\lambda_{\mathrm{hbond}}$ and is only active when polar hydrogens are represented explicitly, since donor atoms only exist as such under that representation.

\paragraph{Total training objective.}
The full loss is the sum of all active terms,
\begin{align}
\label{eq:total_loss}
\mathcal{L} =\;
&\mathcal{L}_{\mathrm{TPXY}}
+ \mathcal{L}_{\mathrm{SLE}}
+ \lambda_{\mathrm{act}}\,\mathcal{L}_{\mathrm{ACT}}
+ \lambda_{\mathrm{tpx}}\,\mathcal{L}_{\mathrm{TPX}}
+ \lambda_{\mathrm{he}}\,\mathcal{L}_{\mathrm{HE}} \notag \\
&+ \lambda_{\mathrm{lle}}\,\mathcal{L}_{\mathrm{LLE}}
+ \lambda_{\mathrm{H}}\,\mathcal{L}_{\mathrm{H}}
+ \lambda_{\mathrm{geom}}\,\mathcal{L}_{\mathrm{geom}}
+ \lambda_{\mathrm{hbond}}\,\mathcal{L}_{\mathrm{hbond}},
\end{align}
with $\lambda_{\mathrm{act}} = 0.5$, $\lambda_{\mathrm{tpx}} = 1.0$, $\lambda_{\mathrm{he}} = 0.5$, $\lambda_{\mathrm{lle}} = 2.0$, $\lambda_{\mathrm{H}} = 0.05$, $\lambda_{\mathrm{geom}} = 0.1$, and $\lambda_{\mathrm{hbond}} = 1.0$. Only terms for which the mini-batch contains data points contribute; a term without active data points contributes zero and produces no gradient. We evaluate the two regularizers separately on the data points of each data type except LLE present in the mini-batch and add their means, so they enter Eq.~\eqref{eq:total_loss} once per data type present.

\paragraph{Optimization.}
The model is trained for 100 epochs with AdamW at a peak learning rate of $5\times 10^{-4}$ and weight decay of $10^{-4}$, following a one-cycle schedule that raises the learning rate over the first $30\%$ of the training steps and anneals it over the remainder (PyTorch \texttt{OneCycleLR} with default settings otherwise, stepped per mini-batch). The batch size is $512$, gradients are clipped to a norm of $5$, and the random seed is $42$. The model is implemented in PyTorch~\cite{Paszke2019} and PyTorch Geometric~\cite{Fey2019}. The ten models of the cross-validation and the full-data model for the multi-component results share this configuration. Throughout this work, we report the final checkpoint rather than the one with the lowest validation loss, so that every model is trained for the same fixed schedule and the validation data are used only for monitoring.

\subsection{Evaluation}
\label{sec:evaluation}
For every data type, the absolute errors of all data points of a system are averaged to its system-wise mean absolute error $\mathrm{MAE}_\mathrm{sys}$, and we report the median, the mean, and the interquartile range of $\mathrm{MAE}_\mathrm{sys}$ over the systems; relative changes between models refer to the median. The errors are taken in $\ln\gamma_i$ for TPXY, ACT, and SLE, in $\ln p$ for TPX, in the mole fractions of all coexisting phases for LLE, and in $h^\mathrm{E}$ (in kJ\,mol$^{-1}$) for HE. Because mod.~UNIFAC does not apply to all systems (missing group decompositions or interaction parameters), the comparisons in the main text are restricted to the systems that all three models can describe; for LLE, additionally restricted to the systems for which all three models predict a phase split for at least one data point. Since every LLE data point corresponds to a measured phase split, this criterion is itself a test of the models, which they pass for a similar share of systems (Supplementary Note~\ref{SI-si:note_benchmark_all}, Supplementary Table~\ref{SI-si:tab_lle_split}). Results on all systems that COGNAC and HANNA describe are given in Extended Data Figs.~\ref{ed:fig_benchmark_all} and~\ref{ed:fig_quaternary}.

HANNA was retrained with its published architecture and hyperparameters on the same data as COGNAC, including the SLE data: one model per fold for the binary cross-validation, and one model on the nine training folds for the multi-component mixtures, with the tenth fold as the validation set that the published HANNA procedure requires and that COGNAC uses only for monitoring, since we report the final checkpoint. The multi-component data were never used for training, but parts of their binary subsystems are contained in the binary training data. Mod.~UNIFAC was used as published; its parameters were fitted to DDB data that include many of the benchmark systems, among them ternary mixtures.

\section*{Data availability}
The experimental data were licensed from the Dortmund Data Bank~\cite{DDB2026} and cannot be redistributed. The COSMO cavity data are from the CHAOS database~\cite{Gond2026} and available on~\href{https://doi.org/10.5281/zenodo.18701299}{Zenodo}. 

\section*{Code availability}
The COGNAC code and the COGNAC and COGNAC-IMP models trained on the full binary are available at \url{https://github.com/marco-hoffmann/COGNAC} under the MIT license. An interactive interface that predicts activity coefficients and phase equilibria and returns parameter tables, an application programming interface, and a Model Context Protocol (MCP)~\cite{MCP2024} connector for applications based on large language models are provided at \href{https://ml-prop.mv.rptu.de}{MLPROP}. Additionally, a Julia implementation of both COGNAC models is available in the Julia package \href{https://se-schmitt.github.io/MLThermoProperties.jl/dev/}{\texttt{MLThermoProperties.jl}}. The code for the batch distillation simulations is provided in \url{https://github.com/se-schmitt/BatchDistillation.jl}.

\section*{Acknowledgements}
The authors gratefully acknowledge financial support by the German Research Foundation (DFG) through the Emmy-Noether program (grant number 528649696), Priority Program SPP 2331 “Machine Learning in Chemical Engineering” (grant number 553345933), Priority Program SPP 2363 “Molecular Machine Learning” (grant number 497201843), and Research Training Group GRK 2908 “WERA” (grant number 503479768), and Germany's Federal Ministry of Economic Affairs and Energy within the project DiKey. Model training and evaluation were carried out on the high-performance computer Elwetritsch at RPTU (grant RPTU-MLVT).

\section*{Author contributions}
M.H.: Conceptualization; Methodology; Software; Validation; Formal analysis; Investigation; Data curation; Writing -- original draft; Visualization. S.S.: Methodology; Software; Formal analysis; Investigation; Writing -- original draft. H.H.: Conceptualization; Writing -- review \& editing; Supervision; Project administration; Funding acquisition. F.J.: Conceptualization; Methodology; Writing -- review \& editing; Supervision; Project administration; Funding acquisition.

\section*{Competing interests}
The authors declare no competing interests.

\bibliographystyle{unsrt}
\bibliography{references}

\clearpage
\renewcommand{\figurename}{Extended Data Fig.}
\renewcommand{\tablename}{Extended Data Table}
\setcounter{figure}{0}
\setcounter{table}{0}

\begin{figure}[p]
\centering
\includegraphics[width=\linewidth]{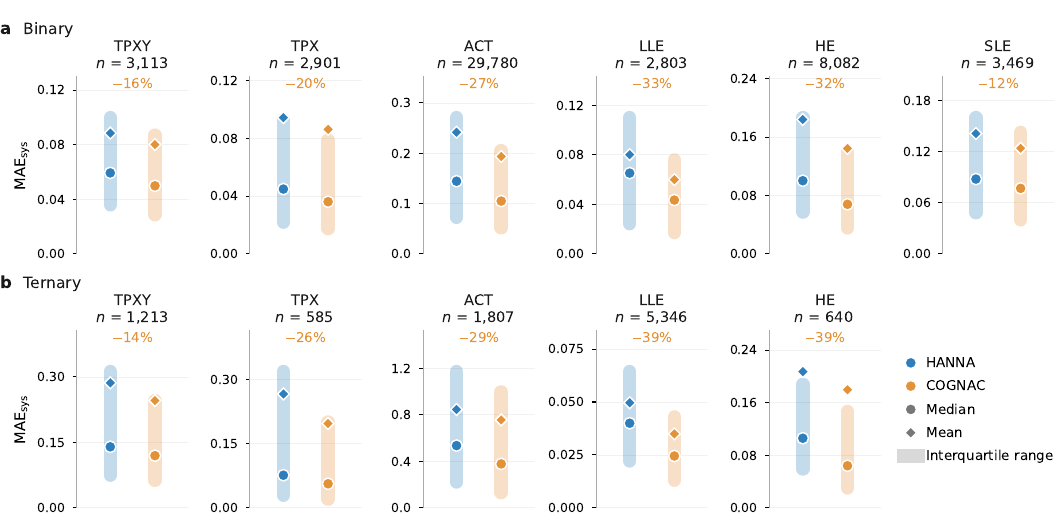}
\caption{Benchmark of COGNAC against HANNA.
\textbf{a}, System-wise mean absolute errors $\mathrm{MAE}_\mathrm{sys}$ for the binary data, resolved by data type; each system is predicted by the COGNAC and HANNA models on the ten-fold cross-validation set it did not see during training.
Circles denote the median, diamonds the mean, and shaded bars the interquartile range of the distribution over systems.
$n$ is the number of systems in each panel. For LLE, a system is included if both models predict a phase split for at least one of its data points. The percentage above each panel shows the relative change in the median $\mathrm{MAE}_\mathrm{sys}$ of COGNAC relative to HANNA; negative values indicate that COGNAC is more accurate. Both models were trained on the same binary data set. The systems outside the coverage of mod.~UNIFAC, included here but not in Fig.~\ref{fig:benchmark_boxplots}, contain components that mod.~UNIFAC cannot decompose or for which interaction parameters are missing; many of them are ionic liquids. \textbf{b} Same comparison for the ternary data set, predicted with the full-data models of COGNAC and HANNA.}
\label{ed:fig_benchmark_all}
\end{figure}

\clearpage
\begin{figure}[p]
\centering
\includegraphics[width=0.667\linewidth]{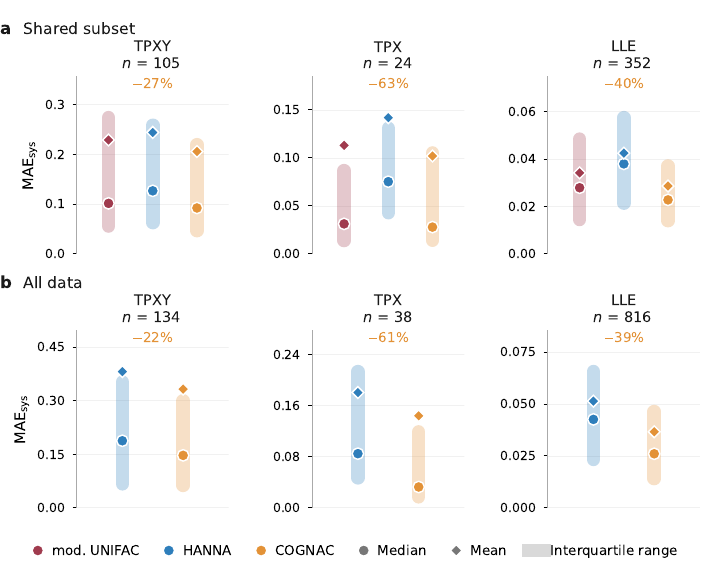}
\caption{Benchmark of COGNAC against mod.~UNIFAC and HANNA on the quaternary data set, analogous to Fig.~\ref{fig:ternary_boxplot}. The data set comprises 21,221 data points for 1,032 systems: 3,750 TPXY data points (134 systems), 1,113 TPX data points (38 systems), and 16,358 LLE data points (888 systems, including 234 three-liquid data points). COGNAC and HANNA were trained only on binary data, so every quaternary prediction extrapolates the number of components; mod.~UNIFAC was used as published and is the only model whose parameters were also fitted to multi-component data. LLE predictions use the same phase-equilibrium procedure as for the ternary mixtures, with the four-component composition space discretized into 6,545 points. System-wise mean absolute errors $\mathrm{MAE}_\mathrm{sys}$ of the logarithmic activity coefficients (TPXY), the logarithmic pressure (TPX), and the liquid mole fractions (LLE, all phases and components). \textbf{a}, Shared subset: systems that all three models can describe; for LLE, only systems for which all three models predict a phase split are included. Relative to mod.~UNIFAC, COGNAC reduces the median $\mathrm{MAE}_\mathrm{sys}$ by 9\% (TPXY), 11\% (TPX), and 18\% (LLE). \textbf{b}, Full data set: all systems that COGNAC and HANNA both describe, without restricting the comparison to the coverage of mod.~UNIFAC. Circles denote the median, diamonds the mean, and shaded bars the interquartile range of the distribution over systems; $n$ is the number of systems. The percentage above each panel shows the relative change in the median $\mathrm{MAE}_\mathrm{sys}$ of COGNAC relative to HANNA; negative values indicate that COGNAC is more accurate. The number of TPX systems is small, so these values should be read with caution.}
\label{ed:fig_quaternary}
\end{figure}

\clearpage
\begin{figure}[p]
\centering
\includegraphics[width=0.9\linewidth]{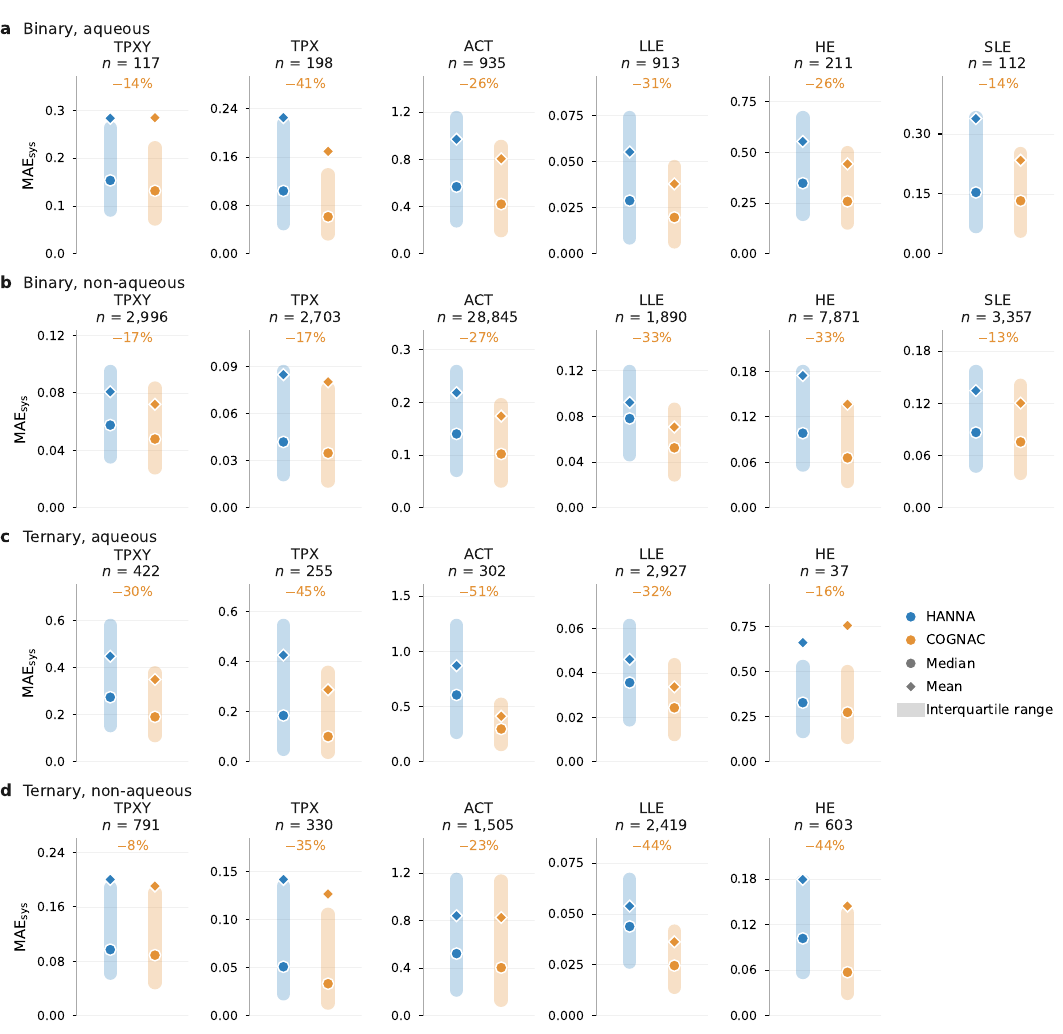}
\caption{Benchmark of HANNA and COGNAC for aqueous and non-aqueous mixtures. System-wise mean absolute errors $\mathrm{MAE}_\mathrm{sys}$ for binary (\textbf{a}, \textbf{b}) and ternary (\textbf{c}, \textbf{d}) mixtures that contain water (\textbf{a}, \textbf{c}) or not (\textbf{b}, \textbf{d}), on all systems that both models describe. Binary: out-of-fold predictions of the ten-fold cross-validation; ternary: full-data models. The data points are those of Extended Data Fig.~\ref{ed:fig_benchmark_all}, split by whether a component is water. Circles denote the median, diamonds the mean, and shaded bars the interquartile range across systems; $n$ is the number of systems. The percentage above each panel shows the relative change in the median $\mathrm{MAE}_\mathrm{sys}$ of COGNAC relative to HANNA; negative values indicate that COGNAC is more accurate. For every data type and both mixture classes, COGNAC gives the lower median $\mathrm{MAE}_\mathrm{sys}$. For the binary mixtures, the relative reduction is similar with and without water except for TPX; for the ternary mixtures, it is considerably larger for aqueous mixtures in TPXY and ACT, larger in TPX, and smaller in LLE and HE.}
\label{ed:fig_aqueous}
\end{figure}

\clearpage
\begin{figure}[p]
\centering
\includegraphics[width=\textwidth]{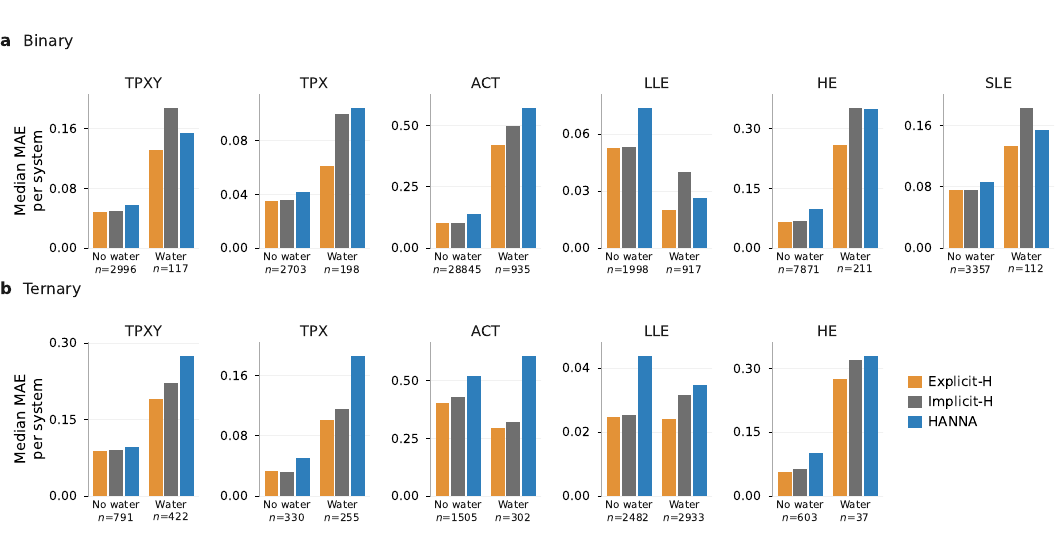}
\caption{Effect of the hydrogen representation on COGNAC accuracy, resolved by mixture and data type. Bars show the median $\mathrm{MAE}_\mathrm{sys}$ of the production model, in which polar hydrogen atoms are explicit graph nodes, and of an ablated model in which all hydrogen atoms are treated implicitly, so that water is a single node. Architecture, training data, loss weights, hyperparameters, and random seed are identical; the only further difference is that the sign regularization of the donor--acceptor interactions, which is defined on the explicit polar hydrogen atoms, has no counterpart in the ablated model, so the comparison measures the combined effect of the explicit hydrogen nodes and this regularizer. We trained the ablated model in every fold of the ten-fold cross-validation and evaluated it like the production model: binary mixtures on out-of-fold predictions, and ternary mixtures with the full-data model, on all systems in these data sets that both models describe; $n$ is the number of systems. For aqueous mixtures, the implicit treatment increases the error for every data type, most strongly for binary LLE ($+101.8\%$) and TPX ($+62.2\%$); without water, the two representations differ by less than $5\%$ except for ternary ACT and HE. For reference, the bars also show the median $\mathrm{MAE}_\mathrm{sys}$ of HANNA, trained on the same folds. HANNA is scored on all systems for which it yields a prediction; for LLE, it predicts no phase split for some systems, which are excluded (binary: 1,840 of 1,998 non-aqueous and 906 of 917 aqueous systems; ternary: 2,414 of 2,482 and 2,897 of 2,933). Even the ablated model is more accurate than HANNA in 18 of the 22 cells; HANNA is ahead only for binary TPXY, LLE, HE, and SLE of aqueous mixtures.}
\label{ed:hydrogen_ablation}
\end{figure}

\clearpage
\begin{figure}[p]
\centering
\includegraphics[width=\linewidth]{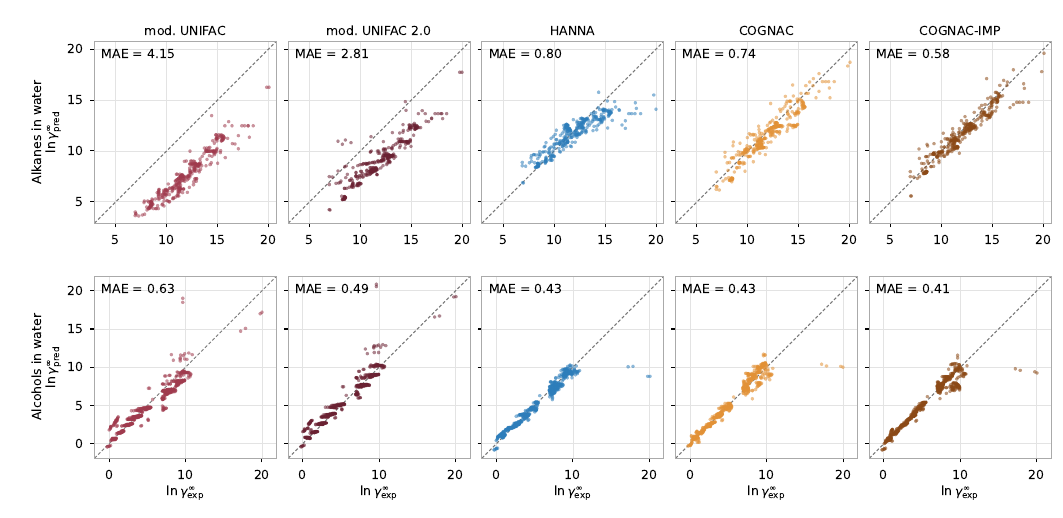}
\caption{Activity coefficients at infinite dilution of alkanes (top) and alcohols (bottom) in water, predicted with mod.~UNIFAC, mod.~UNIFAC 2.0, HANNA, COGNAC, and COGNAC-IMP, against the experimental values: binary ACT data with water as the solvent, with the solutes classified by their DDB component class. COGNAC, COGNAC-IMP, and HANNA: out-of-fold predictions of the ten-fold cross-validation. Mod.~UNIFAC 2.0~\cite{Hayer2025a} is a version of mod.~UNIFAC in which the entire interaction-parameter matrix was fitted simultaneously via matrix completion rather than sequentially; both mod.~UNIFAC versions are used as published and were fitted to DDB data that likely include these data points, which favors them. Only data points for which all models yield a prediction are shown (367 data points of 41 alkane--water and 784 data points of 67 alcohol--water systems). MAE is the mean absolute deviation in $\ln\gamma_i^\infty$. Mod.~UNIFAC underestimates $\ln\gamma_i^\infty$ of every alkane in water; mod.~UNIFAC 2.0 reduces the error but still underestimates $97\%$ of the alkane data points, so the deficiency is not an artifact of the sequential fitting but of the static group decomposition itself. COGNAC, which evaluates the same equations with atom-level parameters, and HANNA, which does not build on a local-composition model, describe both families well.}
\label{ed:fig_aqueous_solute_parity}
\end{figure}

\clearpage
\begin{figure}[p]
\centering
\includegraphics[width=0.5\linewidth]{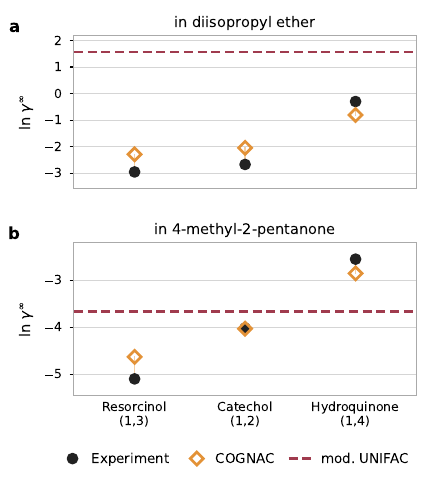}
\caption{Activity coefficients of the three benzenediol isomers at infinite dilution in diisopropyl ether (\textbf{a}) and in 4-methyl-2-pentanone (\textbf{b}) at $298.15\,\mathrm{K}$, predicted with mod.~UNIFAC and COGNAC. In mod.~UNIFAC, the three isomers have identical group counts, so it predicts one value for all benzenediols. Experimental values are shown as filled circles, and COGNAC predictions are shown as open diamonds connected by a thin line to the corresponding measurement. The substitution pattern of each component is given in parentheses. All six COGNAC activity coefficients are out-of-fold predictions; i.e., each was predicted by a cross-validation model that did not see the respective mixture during training.}
\label{ed:fig_degeneracy}
\end{figure}

\clearpage
\begin{figure}[p]
\centering
\includegraphics[width=\linewidth]{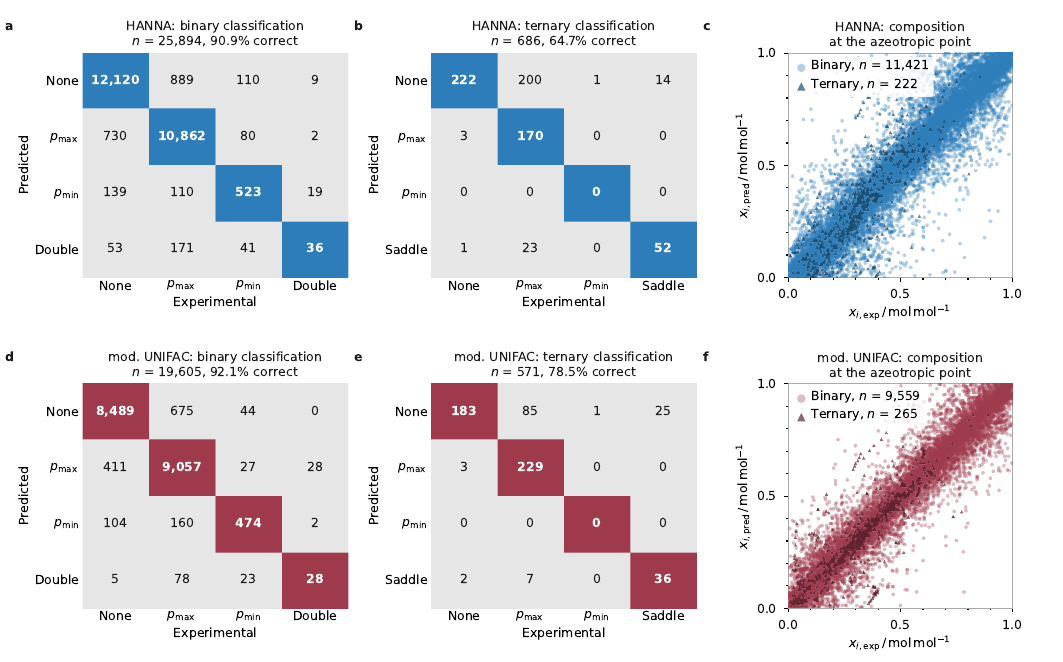}
\caption{Azeotrope predictions of HANNA (\textbf{a}--\textbf{c}) and mod.~UNIFAC (\textbf{d}--\textbf{f}), analogous to Fig.~\ref{fig:azeotropes}a--c. \textbf{a,b,d,e}, Classification matrices for binary (\textbf{a,d}) and ternary (\textbf{b,e}) mixtures, predicted class against experimental class; the diagonal marks correct prediction of azeotrope existence and class. \textbf{c,f}, Predicted against experimental compositions at the azeotropic point, over all components, for the azeotropes whose existence and type were predicted correctly, excluding double azeotropes. Each panel shows the model on all data points for which it yields a prediction, so $n$ is smaller for mod.~UNIFAC; the classification shares compared in Supplementary Note~\ref{SI-si:note_azeotrope_models} are evaluated on the data points that all three models describe.}
\label{ed:fig_azeotrope_models}
\end{figure}

\clearpage
\begin{figure}[p]
\centering
\includegraphics[width=\linewidth]{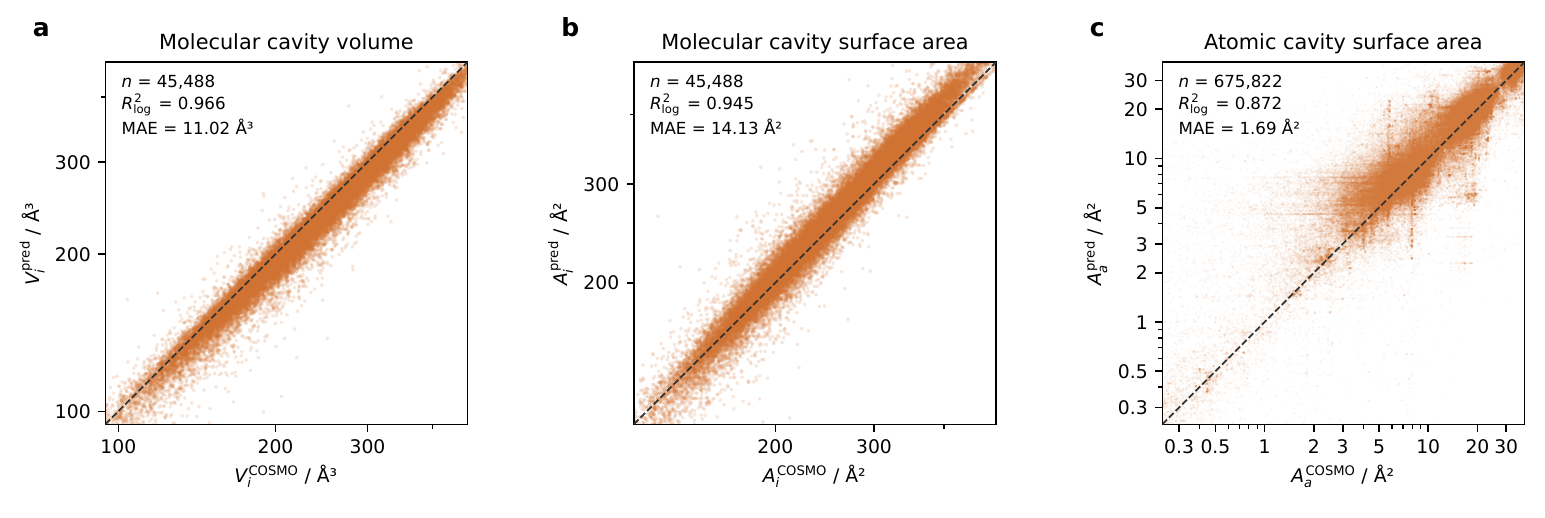}
\caption{COGNAC cavity geometries compared with COSMO references from the CHAOS database~\cite{Gond2026}, predicted by the full-data COGNAC model.
\textbf{a}, molecular volume $V_i$;
\textbf{b}, molecular surface area $A_i$;
and \textbf{c}, atomic surface area $A_a$.
Molecular values are the sums of the atom-level geometry-head outputs, $V_i^{\mathrm{pred}} = \sum_{a\in i} V_a$ and $A_i^{\mathrm{pred}} = \sum_{a\in i} A_a$. The comparison is restricted to the 45,488 CHAOS molecules that occur in no mixture of the thermodynamic data and consist only of elements in the data set; since the COSMO regularizer used in training covers only molecules in the thermodynamic data, none of the compared values entered training. Atom-level areas are available for 45,427 of these molecules (675,822 atoms); carbon-bonded hydrogen areas are folded onto their parent atoms, and polar hydrogens are explicit, consistent with the model's molecular graphs. Superscripts $\mathrm{pred}$ and $\mathrm{COSMO}$ distinguish predictions and reference values; dashed lines indicate parity. The axis limits run from the first percentile of the smaller coordinate in each pair to the 99th percentile of the larger one, retaining about $98\%$ of the points; the reported counts and statistics include points outside the displayed limits, excluding 701 zero-area references from the logarithmic comparison. $R^2_{\log}$ denotes the coefficient of determination against parity in logarithmic space; MAE is calculated in the physical units shown and amounts to 11.02\,\AA$^3$ (molecular volume), 14.13\,\AA$^2$ (molecular surface area), and 1.69\,\AA$^2$ (atomic surface area).}
\label{ed:fig_cosmo_geometry}
\end{figure}

\clearpage
\begin{table}[p]
    \centering
    \caption{Ablation study comparing six COGNAC model variants. The models differ in the implemented temperature dependence (UNIFAC, free form, mod.~UNIFAC) and the closure of the residual activity coefficients (explicit, implicit). Each variant was trained on the ten folds of the cross-validation (Methods) with identical hyperparameters and the same single random seed per fold, so that the variants differ only in the stated switches (the UNIFAC variants also in the combinatorial term, see Supplementary Note~\ref{SI-si:note_ablation}); differences of a few percent are within the variation between independently trained models. Scores are the mean $\mathrm{MAE}_\mathrm{sys}$ over the out-of-fold predictions of all binary systems; as for the comparison of the closures in Supplementary Note~\ref{SI-si:note_imp}, the mean over the systems is used rather than the median of the benchmarks, so that the tail of the error distribution is included. The lowest value per data type among the variants is printed in bold. HANNA denotes the HANNA results of the same ten-fold cross-validation; for LLE, they yield a prediction for 2,653 of the 2,784 systems and are scored on these. COGNAC is the mod.~UNIFAC variant with explicit closure; COGNAC-IMP is the one with implicit closure (Supplementary Note~\ref{SI-si:note_imp}).}
    \label{ed:tab_ablation_study}
    \footnotesize
    \begin{tabular}{ll|cccccc}
    \toprule
       Temperature dependence & Closure & TPXY & TPX & ACT & LLE & SLE & HE \\
      \midrule
      UNIFAC & explicit & $0.083$ & $0.090$ & $0.213$ & $0.066$ & $0.134$ & $0.147$ \\
      UNIFAC & implicit & $0.085$ & $0.092$ & $0.201$ & $0.067$ & $0.129$ & $0.159$ \\
      \midrule
      free form & explicit & $0.078$ & $0.088$ & $0.198$ & $0.062$ & $0.128$ & $0.147$ \\
      free form & implicit & $\mathbf{0.077}$ & $\mathbf{0.085}$ & $0.188$ & $0.060$ & $0.124$ & $0.139$ \\
      \midrule
      mod.~UNIFAC & explicit & $0.080$ & $0.086$ & $0.194$ & $\mathbf{0.059}$ & $0.124$ & $0.145$ \\
      mod.~UNIFAC & implicit & $0.078$ & $\mathbf{0.085}$ & $\mathbf{0.185}$ & $0.060$ & $\mathbf{0.121}$ & $\mathbf{0.138}$ \\
      \midrule
      HANNA & & $0.088$ & $0.094$ & $0.242$ & $0.077$ & $0.141$ & $0.184$ \\
    \bottomrule
    \end{tabular}
\end{table}

\end{document}